\documentclass{article}

\usepackage[square,numbers]{natbib}
\usepackage[preprint]{neurips_2025}

\usepackage[utf8]{inputenc} 
\usepackage[T1]{fontenc}    
\usepackage{hyperref}       
\usepackage{url}            
\usepackage{booktabs}       
\usepackage{amsfonts}       
\usepackage{nicefrac}       
\usepackage{microtype}      
\usepackage{xcolor}         
\usepackage{amsmath}
\usepackage{graphicx}
\usepackage[breakable]{tcolorbox}  
\usepackage{CJKutf8}        
\usepackage{amssymb} 
\usepackage{multirow}       
\usepackage{mathtools}
\usepackage{listings}
\usepackage{subcaption}
\usepackage{caption}

\definecolor{darkblue}{rgb}{0, 0, 0.5}

\newcommand\methodname{\textcolor{black}{\textsc{DataRubrics}}}

\title{Datasheets Aren’t Enough: $\methodname$ for Automated Quality Metrics and Accountability}

\author{
  Genta Indra Winata$^{1,}$\thanks{The authors contributed equally.} \text{ }, David Anugraha$^{2,*}$, Emmy Liu$^{3,*}$, Alham Fikri Aji$^{4,*},$ \\
  \textbf{Shou-Yi Hung$^5$, Aditya Parashar$^1$, Patrick Amadeus Irawan$^6$, Ruochen Zhang$^7$,} \\
  \textbf{Zheng-Xin Yong$^7$, Jan Christian Blaise Cruz$^4$, Niklas Muennighoff$^2$, Seungone Kim$^3$,}\\
  \textbf{Hanyang Zhao$^8$, Sudipta Kar$^9$, Kezia Erina Suryoraharjo$^{5}$, M. Farid Adilazuarda$^4$}, \\
  \textbf{En-Shiun Annie Lee$^{5,10}$, Ayu Purwarianti$^{6}$, Derry Tanti Wijaya$^{11}$, Monojit Choudhury$^{4}$} \\
  $^1$Capital One$\quad$$^2$Stanford University$\quad$$^3$Carnegie Mellon University$\quad$$^4$MBZUAI$\quad$\\
  $^5$University of Toronto$\quad^6$ITB$\quad^7$Brown University$\quad$$^8$Columbia University$\quad$$^{9}$Oracle\\
  $^{10}$Ontario Tech University$\quad^{11}$Monash University \\
}

\begin{document}

\maketitle

\begin{abstract}

High-quality datasets are fundamental to training and evaluating machine learning models, yet their creation—especially with accurate human annotations—remains a significant challenge. Many dataset paper submissions lack originality, diversity, or rigorous quality control, and these shortcomings are often overlooked during peer review. Submissions also frequently omit essential details about dataset construction and properties. While existing tools such as datasheets aim to promote transparency, they are largely descriptive and do not provide standardized, measurable methods for evaluating data quality. Similarly, metadata requirements at conferences promote accountability but are inconsistently enforced. To address these limitations, this position paper advocates for the integration of systematic, rubric-based evaluation metrics into the dataset review process—particularly as submission volumes continue to grow. We also explore scalable, cost-effective methods for synthetic data generation, including dedicated tools and LLM-as-a-judge approaches, to support more efficient evaluation. As a call to action, we introduce $\methodname$, a structured framework for assessing the quality of both human- and model-generated datasets. Leveraging recent advances in LLM-based evaluation, $\methodname$ offers a reproducible, scalable, and actionable solution for dataset quality assessment, enabling both authors and reviewers to uphold higher standards in data-centric research. We also release code to support reproducibility of LLM-based evaluations at \url{https://github.com/datarubrics/datarubrics}.
\end{abstract}

\section{Introduction}

Building high-quality datasets is crucial for effectively developing machine learning (ML) models, as it directly enhances the reproducibility and trustworthiness of the research outcome. However, creating such datasets from scratch is time-consuming and costly, particularly when precise human annotation is required. To address these challenges, there is a growing trend toward generating datasets entirely using Large Language Models (LLMs), offering faster and more cost-effective alternatives. Although this approach provides clear efficiency benefits, it often comes at the cost of diversity, potentially limiting robustness across domains~\cite{anugraha2024proxylm}, as well as originality~\cite{putri2024can} and the rigor of human annotation~\cite{wang2024human}.  Moreover, many LLM-generated datasets lack robust quality assurance and sufficient human oversight, raising concerns about their reliability and downstream utility. These concerns become even more critical when datasets are generated for languages or tasks where models remain empirically weak, such as low-resource languages~\cite{farhansyah2025language,nasution2024chatgpt} or culturally nuanced data~\cite{naous2024having,mihalcea2025ai}. Without proper validation, this can create a vicious cycle, feeding poor-quality data back into LLMs~\cite{shumailov2024ai}. That said, synthetic data is not inherently worse than human-generated data, which can also suffer from biases, errors, and inconsistencies~\cite{veselovsky2023artificial, wu2025style}. Therefore, regardless of the source, mechanisms must be in place to ensure dataset quality.

Efforts such as datasheets and annotation guidelines~\cite{gebru2021datasheets, holland2018dataset, bender2018data, reuel2024betterbench} have been introduced to highlight important quality aspects during dataset creation. While these tools help surface key considerations, they often lack standardized, objective methods for measuring dataset quality. Without clearly defined rubrics or quantifiable metrics, these resources remain general references rather than actionable evaluation frameworks. Consequently, dataset creators lack systematic tools to assess and improve data quality, making it difficult to maintain consistently high standards across projects.

Several major natural language processing (NLP) and ML venues—including the NeurIPS Datasets and Benchmarks Track, Association of Computational Linguistics Rolling Review (ARR), and Language Resources and Evaluation (LREC)—have introduced metadata requirements as part of the submission process to promote transparency and assist reviewers in assessing dataset quality. However, these measures are often inconsistently applied. Authors may provide vague or superficial metadata, and reviewers frequently lack the tools, time, or guidance to interpret this information effectively. As a result, these checks risk becoming mere procedural formalities rather than meaningful quality controls. Moreover, datasheets are seldom used by dataset users, largely due to their open-ended format and absence of structured, quantifiable measures of dataset utility.

While existing datasheets—such as those proposed by~\citet{gebru2021datasheets}—are valuable for highlighting key considerations, they are not designed to be measurable. They rely heavily on open-ended questions without associated assessment criteria, leaving quality assessment up to the subjective interpretation of accompanying papers, if such papers exist at all. Currently, no standard mechanism exists to automate or systematically evaluate dataset quality. Existing rubrics primarily serve an informational role, helping authors and reviewers recognize important dimensions of dataset creation but falling short of enabling objective measurement.

\textbf{In this position paper, we advocate for a more systematic approach to evaluating datasets—moving beyond datasheets and checklists, which are often neither easily decomposable into qualitative and quantitative metrics nor particularly useful for conference paper reviewers.} In particular, we advocate adopting clear, rubric-based evaluation frameworks that enable consistent, reproducible assessments of dataset quality, thereby strengthening both the review process and the overall integrity of dataset-driven research. Several recent studies have explored the use of LLMs for evaluating reward models and benchmarking, including RM-R1~\cite{chen2025rm}, FLAME~\cite{vu2024foundational}, Cloud~\cite{ankner2024critique}, MetaMetrics~\cite{winata2025metametrics}, and R3~\cite{anugraha2025r3}, a paradigm referred to as \textit{LLM-as-a-judge}. Our approach which uses structured rubrics is compatible with both human evaluation as well as LLM-as-a-judge approaches, and could be scaled to improve the paper reviewing process through enriching papers with metadata on dataset quality.

\begin{figure*}[!t]
    \centering
    \includegraphics[width=\textwidth]{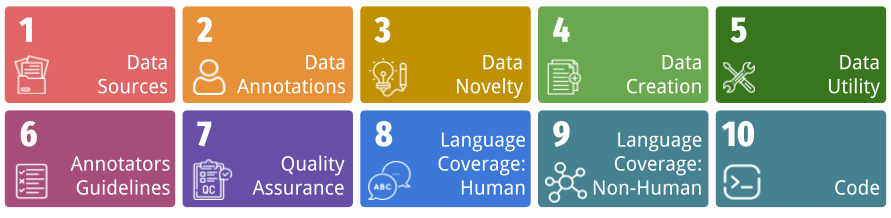}
    \caption{We study the datasheets and checklists in 10 dimensions of data quality.}
    \label{fig:datarubrics}
\end{figure*}


To support our position, we conduct a comprehensive analysis of datasheets and checklists used in leading academic conferences across ML, NLP, computer vision (CV), and speech. Our findings reveal significant gaps in the information provided to assess data quality and enable effective benchmarking—ultimately undermining the reliability of current evaluation practices. We examine the challenges of assessing data quality across multiple dimensions and advocate for a set of key principles and best practices to guide more effective and consistent evaluations. As a call to action, we introduce $\methodname$, a systematic evaluation metric designed to measure the quality of both human- and model-generated datasets. Based on our analysis, we identify ten critical aspects of data quality and develop \methodname{} as a structured, rubric-based framework grounded in quantifiable criteria—such as data source, annotation methods, quality assurance processes, originality, reproducibility, and documentation. We observe that major ML and NLP conferences lack standardized, quantifiable criteria for data quality assessment in their submission checklists and datasheets, a gap that may contribute to the acceptance of low-quality datasets. To investigate this further, we use $\methodname$ to manually annotate 100 papers submitted to the NeurIPS Datasets and Benchmarks Track, followed by a quality assurance process. We also apply LLM-based evaluation using $\methodname$ to papers from a range of top-tier conferences in NLP, CV, ML, and speech, analyzing patterns and trends in data quality over time.

\section{Datasheets Are Not Enough}
\subsection{Standardized, but Lacking Qualitative Evaluation}
In recent years, the research community has introduced datasheets as a tool to promote standardization and raise awareness among authors of dataset papers. These datasheets aim to encourage thoughtful reflection on what constitutes a high-quality dataset. Standardization has been beneficial—not only for authors, but also for reviewers tasked with assessing dataset quality. However, while datasheets enhance transparency and documentation, they do not offer a systematic framework for objectively evaluating dataset quality.

The datasheet template proposed by~\citet{gebru2021datasheets} outlines several key dimensions: (1) motivation for dataset creation, (2) dataset composition, (3) data collection process, (4) data preprocessing, (5) dataset distribution, (6) dataset maintenance, and (7) legal and ethical considerations. Understanding the motivation and composition of a dataset is undeniably valuable, as it sheds light on its intended use, relevance to downstream tasks, size, and structure. However, while these categories enhance transparency, they do not inherently facilitate measurable quality assessment. The framework lacks standardized rubrics or scoring mechanisms that would enable consistent, systematic evaluation across datasets. Moreover, in the era of large-scale pre-trained models, the existing datasheet categories fall short in addressing model-specific considerations that are increasingly important in dataset design. As LLMs are often both the consumers and generators of data, evaluation frameworks must adapt accordingly. A datasheet approach tailored to the unique challenges and evaluation needs of LLM-centric workflows would be more relevant and impactful for modern dataset development.

Another valuable contribution is the “Data Statements for NLP” framework by~\citet{bender2018data}, which emphasizes that not only dataset composition and distribution matter, but also the role and background of annotators—highlighting that sociolinguistic and ethical considerations are often overlooked. Despite this, our analysis of recent dataset papers published at major NLP and ML conferences reveals that reporting on annotator demographics and perspectives remains sparse and inconsistent. Such guidelines exist, but many authors engage only superficially with annotator-related documentation, limiting the transparency and reproducibility of the data annotation process.

\subsection{The State of Academic Conferences}
Datasheet frameworks have been increasingly encouraged by conference organizers to ensure comprehensive documentation of dataset creation processes. In this paper, we examine the policies and initiatives adopted by academic conferences to promote reproducibility and uphold responsible standards for dataset and benchmark submissions. Additionally, we analyze conferences that have not yet adopted or advocated for the use of datasheets or data checklists, highlighting gaps and opportunities for broader standardization across the research community.

\paragraph{NeurIPS Dataset and Benchmark Track.}
Since launching its Datasets and Benchmarks track in 2021, NeurIPS has progressively introduced policies to improve documentation, transparency, and accountability in dataset submissions. Authors have been encouraged to adopt established frameworks such as Datasheets for Datasets~\cite{gebru2021datasheets}, Dataset Nutrition Labels~\cite{holland2018dataset}, Data Statements for NLP~\cite{bender2018data}, and broader accountability frameworks~\cite{hutchinson2021towards}. Submissions are also required to include a public URL to the dataset or benchmark and a clearly defined data license—both essential for ensuring reproducibility and reuse. In 2022, NeurIPS strengthened its licensing policy by requiring authors to explicitly specify a valid data license. That same year, it introduced a new policy mandating the use of structured metadata standards (e.g., schema.org, DCAT) to enhance machine-readability and discoverability. In 2023, the track added a requirement for disclosing any use of large language models (LLMs), a policy that continued into 2024. Looking ahead, the 2025 track introduces two additional requirements: the use of structured metadata and the submission of a formal checklist to improve transparency. Notably, authors must now format metadata according to the Croissant standard~\cite{akhtar2024croissant}, a step toward greater standardization and interoperability. While these evolving policies represent significant progress in promoting high-quality documentation and responsible dataset practices, they still primarily emphasize descriptive metadata. As a result, they provide transparency but fall short of offering evaluative insights—they do not directly measure the quality of the dataset itself. There remains considerable room for improvement, particularly in making the metadata more actionable and useful for reviewers seeking to assess the quality and reliability of a dataset.

\paragraph{ICML and ICLR.} According to Figure~\ref{fig:number_of_papers}, ICML and ICLR have relatively few accepted dataset and benchmark papers compared to other conferences, and they do not require authors to submit any accompanying checklists or datasheets for these submissions.

\paragraph{NLP Conferences.}
In NLP conferences such as *CL and EMNLP, authors are encouraged—but not required—to complete Responsible NLP forms and datasheets when submitting their papers.\footnote{\url{https://aclrollingreview.org/responsibleNLPresearch}.} This practice is particularly emphasized within the ARR system, which functions as the unified submission portal for *CL, EMNLP, and affiliated workshops. These forms are intended to promote transparency and responsible research practices. In contrast, conferences focused on language resources, such as LREC, require authors to disclose dataset information at the time of submission. However, unlike NeurIPS 2025, LREC has not yet mandated the use of structured datasheets or standardized metadata.

\paragraph{CV Conferences.}
To the best of our knowledge, major computer vision conferences such as CVPR, ECCV, and ICCV do not currently enforce or actively advocate for the inclusion of datasheets or checklists as part of supplementary materials. This lack of emphasis suggests a limited focus on systematically measuring and documenting dataset quality within the CV research community.

\paragraph{Speech Conferences.}
The Interspeech conference began implementing a checklist in 2024, with continued use in 2025. This checklist collects detailed information about the models and datasets used, including dataset statistics. Authors are expected to disclose whether the datasets are publicly available or not, promoting transparency and helping readers understand the scope and accessibility of the data. Additionally, the checklist requires authors to report key details such as the languages covered, total audio duration, number of examples, and label distributions. These elements provide valuable insights that enable readers to better assess and compare dataset quality.

\subsection{Limited Coverage}
The absence of both qualitative and quantitative rubrics has led to limited coverage in evaluating dataset quality. As shown in Table~\ref{tab:compare-benchmark}, most existing datasheet frameworks lack sufficient scope. BetterBench~\cite{reuel2024betterbench} represents the most comprehensive effort to date, primarily due to its use of rubrics that enable more structured and quantifiable assessments. However, it still
falls short in evaluating important dimensions such as data novelty and language diversity, and its
evaluations are conducted manually rather than automatically, limiting scalability. Therefore, we advocate developing a more robust and automated set of metrics to enable comprehensive, systematic, and scalable evaluation of dataset quality.

Another notable limitation of BetterBench is its lack of support for non-human language coverage—such as scientific datasets using structured representations like molecular schemas (e.g., proteins) or neurosymbolic formats that rely on domain-specific grammars. In contrast, $\methodname$ explicitly incorporates non-human language dimensions, making it extensible to a broader range of knowledge domains beyond human-centric language data.

\begin{table*}[!t]
\centering
\caption{Coverage of several popular dataset documentation schemas, including \methodname.}
\label{tab:compare-benchmark}
\resizebox{\textwidth}{!}{
    \begin{tabular}{l|ccccc|c|c|cc|c|c}
    \toprule
    & \multicolumn{5}{c|}{\textbf{Data}} & \textbf{Annotators} & \textbf{Quality} & \multicolumn{2}{c|}{\textbf{Languages Coverage}} & \textbf{Code} & \textbf{Rubrics} \\ 
    & Sources & Annotation & Novelty & Creation & Utility & Guidelines & Assurance & Human & Non-Human & Provided \\
    \midrule
    Datasheets~\cite{gebru2021datasheets} & $\checkmark$ &  &  & $\checkmark$ & $\checkmark$ &  & &  &  &  &\\
    Data Nutrition~\cite{holland2018dataset} & $\checkmark$ & & & & & & & & & & \\
    Data Statements~\cite{bender2018data} & $\checkmark$ & $\checkmark$ & &  & &  &  & $\checkmark$ & & & \\
    BetterBench~\cite{reuel2024betterbench} & $\checkmark$ & $\checkmark$ &  & $\checkmark$ & $\checkmark$  & $\checkmark$ & $\checkmark$ &  &  & $\checkmark$ & $\checkmark$\\ \midrule
    $\methodname$ & $\checkmark$ & $\checkmark$ & $\checkmark$ & $\checkmark$ & $\checkmark$ & $\checkmark$ & $\checkmark$ & $\checkmark$ & $\checkmark$& $\checkmark$ & $\checkmark$ \\
    \bottomrule
    \end{tabular}
}
\vspace{-3mm}
\end{table*}

\subsection{Rubrics and Scoring Matter}
The checklists currently used in conferences are not based on rigorous rubrics. They typically rely on binary responses such as “yes,” “no,” or “N/A,” which lack explainability and fail to capture the complex, multi-faceted characteristics of datasets. For example, a dataset might be annotated both by humans and with assistance from LLMs, but such hybrid methods are difficult to identify using existing checklists or most datasheets—unless a more detailed framework like BetterBench is employed. However, some of BetterBench’s scoring systems may not generalize well across all contexts. For instance, a dataset accepted at a workshop is not necessarily of lower quality than one presented at a major conference, and domain-specific datasets may not fit uniform metric standards. To address these challenges, we propose a more general set of attributes and dimensions for assessing datasets and benchmark papers. These are designed to be robust across domains and better suited for automated evaluation, including LLM-as-a-judge scenarios.

\section{Automating Dataset Quality Assessment: A Call to Action}
Initially, Datasheets~\cite{gebru2021datasheets} were designed not as an automated evaluation framework, but to encourage dataset creators to thoughtfully reflect on the processes of creating, distributing, and maintaining a dataset. However, with the growing volume of conference submissions (see Figure~\ref{fig:number_of_papers}) and varying reviewer expertise, providing a clear automated evaluation could help reviewers quickly assess dataset utility and quality. While this rubric is not meant to replace human expert judgment, it can guide reviewers to key aspects, such as the extent of synthetic data usage, dataset modalities, and quality control measures implemented by the authors.

\subsection{Reviewers May Overlook or Struggle to Assess the Quality of the Datasets}

Anecdotal stories have emerged of papers being rejected due to having too few languages or small dataset sizes. With the rise of LLMs and the increasing use of synthetic data generation, reviewers may overlook the fact that dataset quality arguably matters more than quantity, and should avoid judging datasets solely based on size. In many cases, a small dataset with few languages may be sufficient to prove a point based on statistical power analysis, or to serve the purpose outlined by the authors. Conversely, a large dataset with many languages may be \textit{insufficient} for the purposes proposed by the authors. A data resource must be judged clearly by its intended purpose, and whether its construction is sufficient to serve as a resource for that purpose.

Without proper quality assessment, reviewers may focus on more visible aspects, such as dataset size or language coverage, rather than critically evaluating the data itself. This issue is even more pronounced when reviewers do not speak the target language, making it difficult to directly assess data quality. Additionally, it is important to note that synthetic data is not inherently low-quality, just as human-annotated data is not automatically high-quality. A clearer and more standardized quality rubric would help reviewers evaluate data-centric work more fairly, beyond surface-level numbers such as data size or language coverage.

\subsection{Surge in Paper Submissions}

\begin{figure*}[!t]
    \centering
    \includegraphics[width=\textwidth]{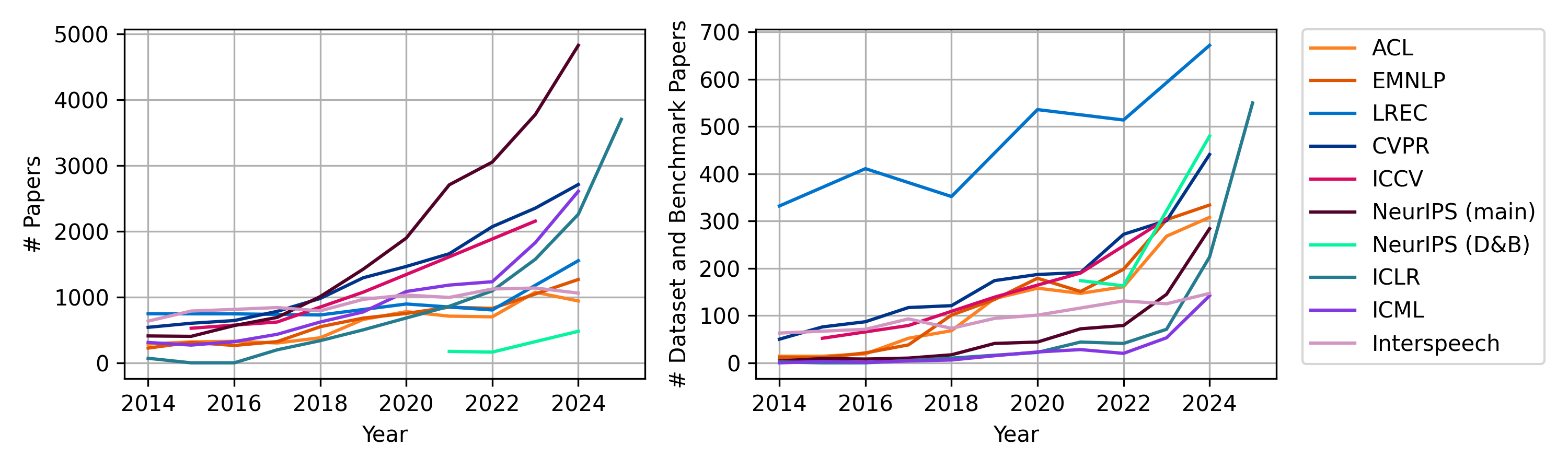}
    \caption{Comparison of yearly trends: all accepted papers \textbf{(left)} and dataset or benchmark papers \textbf{(right)}. The number of dataset and benchmark papers is determined by classifying paper titles and abstracts. As of the time of writing, only the ICLR 2025 proceedings had been published and are therefore the only 2025 data included.}
    \label{fig:number_of_papers}
    \vspace{-2mm}
\end{figure*}

In recent years, AI conferences have seen a massive surge in submissions (see \autoref{fig:number_of_papers}). As the number of submissions grows, so does the number of required reviews, leading to an increased reviewer pool and load and, in some cases, more relaxed reviewer eligibility criteria. With this rising review burden, maintaining review quality becomes increasingly challenging. To support reviewers, especially in evaluating dataset-related work, more comprehensive and up-to-date data cards that highlight dataset quality are needed. Additionally, automated tools that assist in the review process by providing proxies for dataset quality could be valuable, helping reviewers make more informed assessments. AI-assisted reviews for data-related work are helpful, given the increasing scale of paper submissions. This tool can be used to quickly highlight data quality issues, including references to the relevant parts of the article for the reviewer to check. 

\section{DataRubrics}
\label{category_defn}

We have discussed the limitations of existing datasheets and checklists used in conferences and strongly advocate for improvements. Specifically, we propose developing new automated rubrics that cover a broader range of data quality dimensions not addressed by current methods. In this section, we present our proposal in detail, focusing on three key concepts to aid reviewers and users in evaluating datasets: (1) Data composition and creation, (2) Quality assurance and reproducibility, and (3) Novelty and differentiation in terms of utility. We summarize our framework’s components in Figure~\ref{fig:datarubrics} and define these facets below.

\subsection{Dimensions}
We define 10 distinct dimensions to capture the key aspects necessary for evaluating data quality.

\paragraph{(1) Data Sources.}
It is important to specify the original data source—whether the content is human-written or machine-generated. This distinction has become increasingly relevant with the rise of AI-generated content, which facilitates the creation of large-scale synthetic datasets. Documenting the data source helps ensure that datasets are used appropriately and in line with their intended purpose, particularly when determining whether machine-generated or human-written data is more suitable. 
\paragraph{(2) Data Annotators.}
Clear documentation of annotators is essential, as human-annotated labels do not inherently guarantee high quality. Inadequate annotation protocols or reliance on non-experts can result in low-quality data. It is important to provide information about the annotators’ demographics and their domain-specific expertise. For instance, native speakers or individuals familiar with local cultures are critical for multilingual or culturally nuanced data, while subject-matter experts are necessary for scientific benchmarks. Similarly, machine-generated data may reflect the biases of the underlying models, which could be problematic—especially if the dataset is intended to fill gaps in existing AI capabilities, such as those involving low-resource languages. Therefore, those involved in dataset construction must understand the limitations of the models used. \paragraph{(3) Data Novelty.}
Data work should also clearly indicate the novelty of the dataset, whether it is constructed from scratch, translated from existing data, derived from other datasets, or simply collected and collated from various sources. This information is important because it provides context on the originality, potential biases, and limitations of the dataset. For instance, translated or derived datasets may carry over biases or assumptions from the source material, while collated datasets might introduce inconsistencies due to differing annotation standards and data qualities. 
\paragraph{(4) Data Creation.}
This metric assesses the transparency and thoroughness of the dataset creation documentation—key factors for ensuring reproducibility, facilitating ethical evaluation, and supporting downstream usability.
\paragraph{(5) Task Utility.}
This metric evaluates how the dataset is utilized within the ML pipeline. Understanding its role helps clarify the dataset’s purpose, relevance, and integration into model development and evaluation workflows. \paragraph{(6) Annotation Guidelines.}
Annotation guidelines are essential for aligning annotators and helping readers understand dataset creation. Machine-generated data also requires annotation guidelines, typically describing how the data or labels were produced, including specific prompts used. These guidelines should provide clear instructions and well-defined rubrics for each category to minimize disagreement. Additionally, annotation examples are also important to enhance annotation quality.
\paragraph{(7) Quality Assurance.} Beyond construction, validation is a vital step in improving dataset quality. Information about who performed the quality assurance—whether an expert or a machine—should be provided. Additionally, a clear guideline outlining how the quality assurance was conducted is necessary, including some notation on the quality itself, for example via annotators' agreement. Machine-generated data may require documentation on how well the machine performs the annotation task. This can be supported by a performance report on similar tasks using third-party benchmarks or by showing correlation with high-quality human annotations. This information helps to ensure that the machine-generated data is aligned with the expectations of human annotators. 
\paragraph{(8) Human Language Coverage.} With the growing focus on NLP research beyond English, it is increasingly important to consider language coverage in dataset development. We recognize that data are not always sourced from English, but also from a wide range of non-English languages.
\paragraph{(9) Non-Human Language Coverage.}
Not all data originates from human languages—some datasets are based on abstract, structured, or symbolic representations, such as those found in scientific or formal languages. Therefore, we also track and include information on non-human languages in our analysis. 
\paragraph{(10) Code.}
This metric pertains to whether the code used for constructing the dataset is made publicly available or not for reproducibility.

\subsection{Rubric-Based Design}
\paragraph{Multi-label with Reasoning and Reference.}
To enable more precise and meaningful evaluation, our rubrics go beyond simple numerical scores by providing clear criteria to distinguish different labels. A key design goal is support for multi-label classification. For example, a dataset may combine human-annotated data with synthetic data generated by LLMs, requiring metrics that accurately capture these multifaceted characteristics. We also emphasize the importance of reasoning and evidence in annotation: since verifying label correctness can be difficult without justification, annotators are required to provide explanations along with references to specific sections of the paper. This improves transparency and enables more verifiable evaluations.

\paragraph{Schema and Structured Decoding.}
Given the rubric’s many dimensions, manual assessment by authors, annotators, or reviewers can be overwhelming. To address this and enable scalable LLM-assisted evaluation, we design the rubric to be both human-readable and machine-interpretable. Specifically, we provide a structured schema that guides LLMs during generation via constrained structured decoding. This ensures consistent, rubric-aligned outputs while making evaluation more efficient and scalable. Details about the schema can be found in Appendix~\ref{appendix:schema}.

\begin{figure*}[!t]
    \centering
    \includegraphics[width=0.9\textwidth]{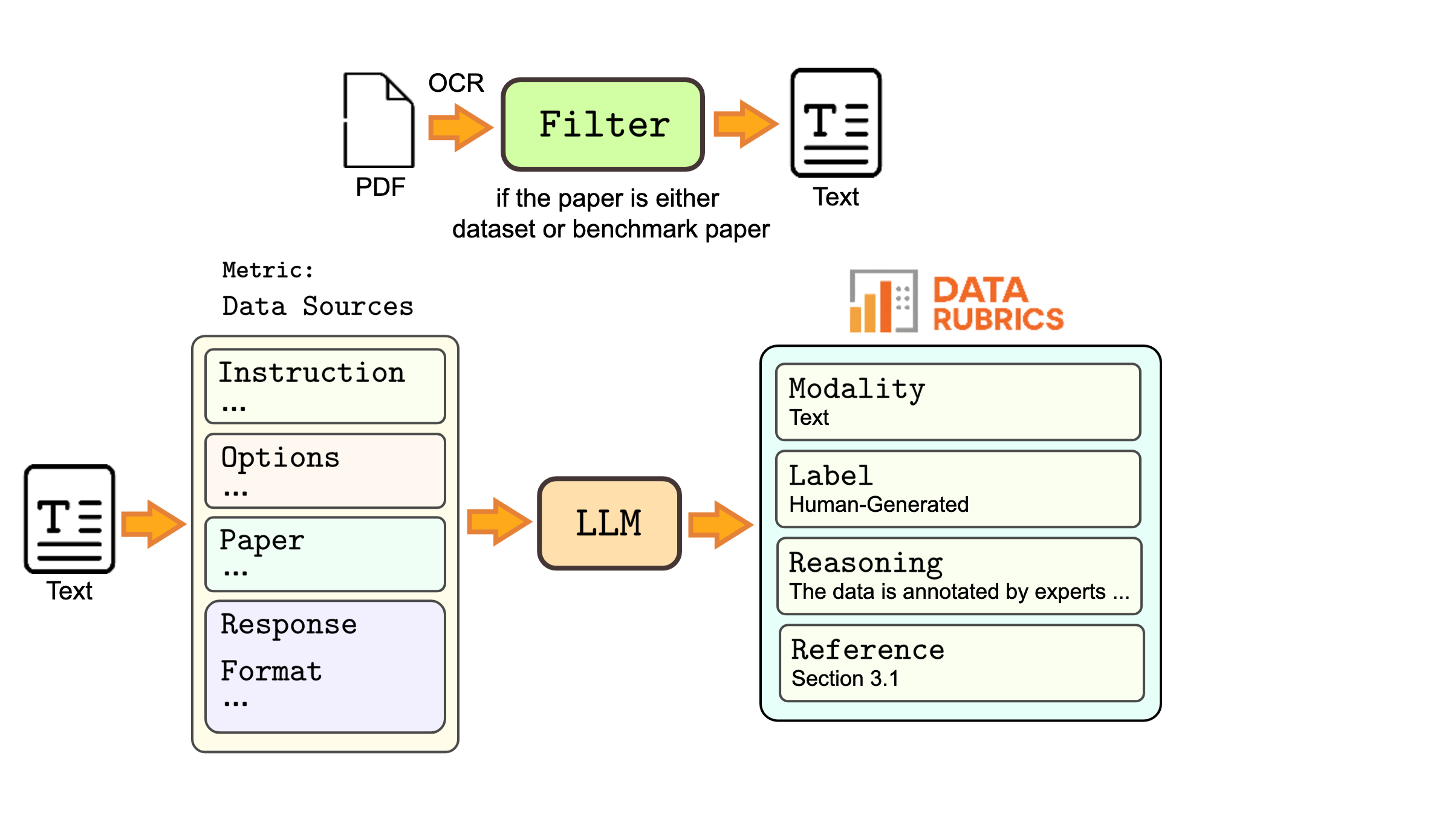}
    \caption{Inference pipeline using DataRubrics. We use OCR to extract text from PDFs, filter papers by title and abstract to select dataset or benchmark papers, then use an LLM to predict labels, reasoning, and references according to the categories in $\S\ref{category_defn}$.}
    \label{fig:example}
    \vspace{-2mm}
\end{figure*}

\section{Resources and Evaluation}

As shown in Figure~\ref{fig:example}, our dataset collection involves several key steps: high-quality text extraction using OCR, filtering with a reward model, and automatic evaluation through LLM-as-a-judge. We describe each step in detail below:

\subsection{Dataset Collection}

We collect text and metadata, including titles and abstracts, from papers published at major conferences across ML/AI (i.e., NeurIPS, ICLR, ICML), NLP (i.e., ACL, EMNLP, LREC), CV (i.e., CVPR), and speech processing (i.e., Interspeech), spanning the years 2021 to 2024, since at the time of writing, not all proceedings from 2025 have been released.

\paragraph{Filtering Using a Reward Model.}

To identify relevant papers, we first narrow the corpus using titles and abstracts to isolate works potentially related to novel datasets or benchmarks. We employ a generative reward model, R3-Qwen3-14B-4k~\cite{anugraha2025r3}, which performs rubric-guided reasoning and reward-based classification. This model helps detect papers that (i) introduce new datasets, (ii) modify or extend existing datasets, or (iii) conduct benchmarking on existing datasets. The model is prompted using a structured template described in Section~\ref{section:appendix-benchmark-classifier-template} of the Appendix. This filtering process yields a substantial set of relevant papers, as illustrated in Figure~\ref{fig:number_of_papers}.

\paragraph{High-Quality Text Extraction via OCR.}
For the filtered papers, standard PDF text extraction methods often yield noisy or incomplete results due to formatting inconsistencies, embedded figures or tables, and anonymization artifacts in under-review manuscripts. To ensure high-quality and structured text extraction, we apply a robust Optical Character Recognition (OCR) model, OlmOCR~\cite{poznanski2025olmocr}, specifically designed for academic document parsing.

\paragraph{Sampling.}
In our study, we limit the analysis to a maximum of 100 papers per year for each conference category (ARR, ICML, ICLR, NeurIPS Datasets and Benchmarks Track, CVPR, Interspeech, and LREC), selected through random sampling. For the human evaluation, we specifically focus on 100 papers from the NeurIPS Datasets and Benchmarks Track, sampling data only from the years 2022 to 2024.

\begin{figure}[!ht]
    \centering
    \begin{subfigure}[b]{\textwidth}
        \includegraphics[width=\textwidth]{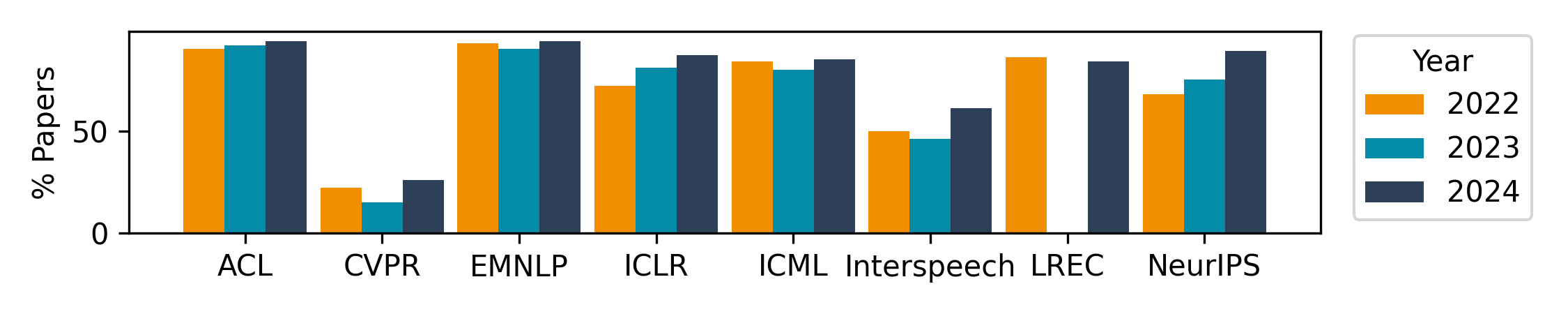}
        \caption{Percentage of papers with guidelines for data collection.}
        \label{fig:sub1}
    \end{subfigure}
    \hfill
    \begin{subfigure}[b]{\textwidth}
        \includegraphics[width=\textwidth]{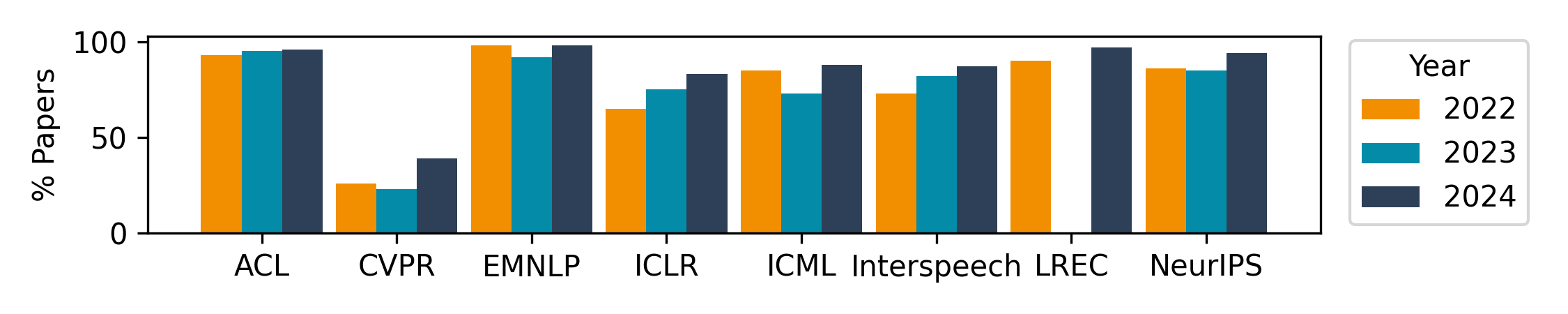}
        \caption{Percentage of papers with quality assurance.}
        \label{fig:sub2}
    \end{subfigure}
    \caption{Paper statistics across conferences. Only the 2022 and 2024 editions of LREC are included, as it is held biennially.}
    \label{fig:paper-statictics-data-quality}
    \vspace{-2mm}
\end{figure}

\subsection{Automatic Evaluation via LLM-as-a-judge}
With high-quality paper texts in hand, we perform automatic evaluation using LLM-as-a-judge, following the $\methodname$ methodology. The models are prompted using a structured template described in Section~\ref{appendix:schema} of the Appendix. We evaluate responses using a proprietary model GPT-4.1-mini.

\subsection{Manual Evaluation}
To evaluate the usefulness of the metric and the assessment pipeline, we engaged domain experts to perform rubric-based annotations. Each paper was assigned to one expert with specialization in ML, NLP, CV, or Speech, who annotated the relevant aspects covered by our metrics. This was followed by a quality assurance step, in which a second expert reviewed the annotations to identify and correct any issues. The human evaluation was conducted on a sample of 100 NeurIPS papers from 2022 to 2024, comprising both accepted and rejected submissions from the Datasets and Benchmarks Track.




\subsection{Analysis}

We analyze annotations from both human annotators and AI model-based evaluations to assess the effectiveness of the rubrics and to examine trends in papers over the past several years.

\subsubsection{Trends in Dataset Papers Across Academic Conferences}

\paragraph{Data Annotation and Quality Assurance.} Figure~\ref{fig:paper-statictics-data-quality} illustrates trends in dataset papers across various academic conferences from 2022 to 2024. Several conferences show an upward trend in the inclusion of guidelines and quality assurance practices, indicating growing awareness of the importance of data standards. However, CVPR consistently exhibits the lowest proportion of such papers, with only a slight improvement observed in 2024. This aligns with our broader findings that CVPR currently lacks strict policies or standardization regarding datasheets and data checklists. We advocate for CVPR to adopt similar initiatives to those introduced by conferences like NeurIPS.

\begin{figure}[!ht]
    \centering
    \includegraphics[width=\textwidth]{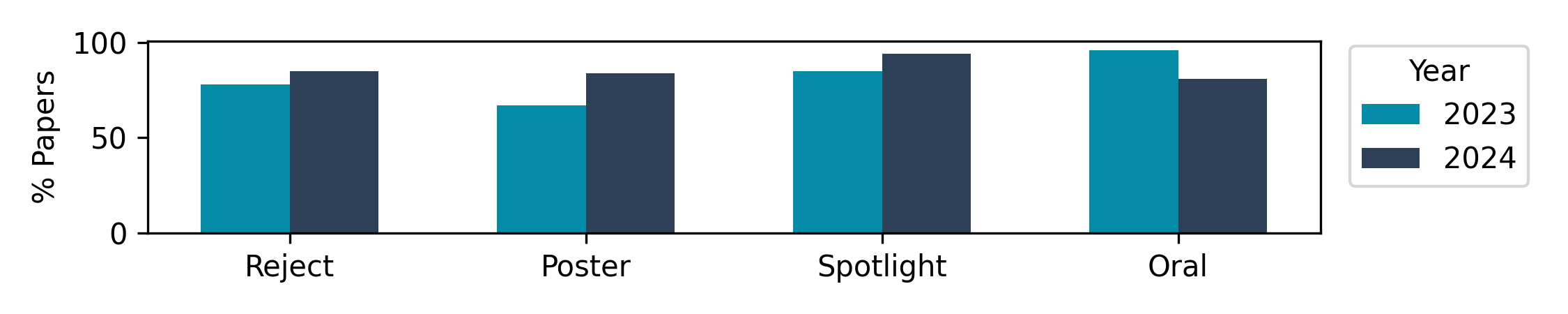}
    \caption{Percentage of papers on NeurIPS with Guidelines on their data collection.}
    \label{fig:percentage-guidelines-data-annotations}
    \vspace{-2mm}
\end{figure}

\paragraph{Accepted vs. Rejected NeurIPS Papers.}
Figure~\ref{fig:percentage-guidelines-data-annotations} shows the percentage of NeurIPS papers across different acceptance and rejection categories. The number of papers in each category is relatively similar, indicating that NeurIPS’ policy on maintaining quality in the dataset and benchmark track is enforced even among rejected submissions. However, there is still room for improvement in raising awareness about the importance of following guidelines during data collection.

\begin{figure}[!ht]
    \centering
    \includegraphics[width=\textwidth]{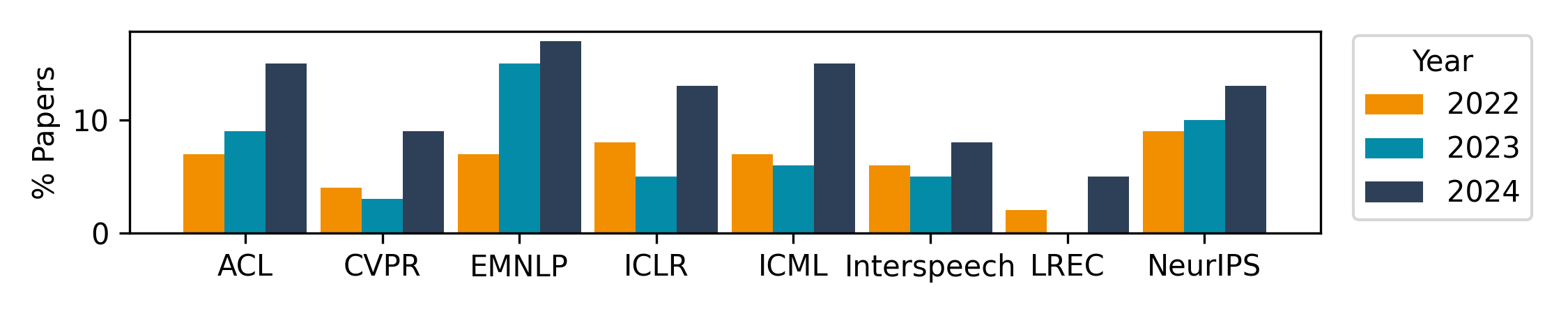}
    \caption{Percentage of papers with proposed new data generated from model.}
    \label{fig:percentage-new-data-from-model}
    \vspace{-2mm}
\end{figure}

\paragraph{Model-generated Data.}
Figure~\ref{fig:percentage-new-data-from-model} shows the trend of papers proposing new data generated by models. There is a clear and consistent increase in the percentage of such papers over time across all conferences.

\subsubsection{Automatic Evaluation vs. Human Evaluation}
\begin{figure*}[!ht]
    \centering
    \includegraphics[width=0.75\textwidth]{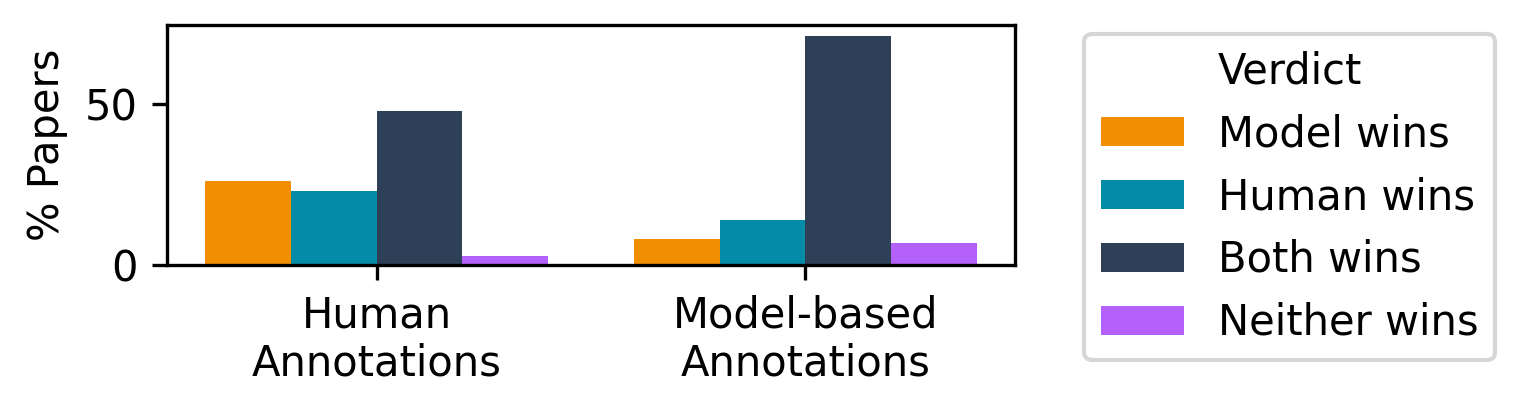}
    \caption{Comparison of human–LLM annotation agreement.}
    \label{fig:after-qa}
    \vspace{-2mm}
\end{figure*}

We find that human annotations still contain errors even after undergoing quality assurance (QA) procedures. To better understand the nature of these errors, we reannotate a set of 100 data points using outputs from GPT-4.1 mini, focusing on NeurIPS papers. As shown in Figure~\ref{fig:after-qa}, 26\% of the annotations remain incorrect, despite having passed QA by human annotators tasked with identifying whether a paper includes specific annotations. Our analysis suggests that human annotators frequently overlook fine-grained details, especially when subtle or nuanced aspects of a paper are involved. To investigate these discrepancies further, we compare annotations across multiple perspectives, including both model and human annotation guidelines. This comparison reveals that a substantial portion of errors stem from misclassifying human-written annotations as model-generated ones. Similar patterns of misclassification are also observed in automatic evaluation metrics. These findings indicate that model-assisted annotation could play a valuable role in reducing human error during the review process. Alternatively, authors should strive to report annotations as clearly and accurately as possible to minimize ambiguity.

\section{Conclusion}

Moving forward, we offer several recommendations for the AI community, particularly those developing datasets and benchmarks. First, documenting all aspects of data work—from sources to annotation processes—is crucial. Our findings highlight the importance of such transparency. Second, data quality must be evaluated beyond surface metrics like size or language fluency. While LLMs enable large-scale dataset creation, quality remains paramount. Machine-generated data should be rigorously validated for human alignment and bias, while human annotations demand clear guidelines and expert oversight to reduce error. As dataset submissions grow, maintaining review quality is increasingly difficult. AI-assisted reviewing—where models summarize data quality based on structured rubrics—could ease this burden. We hope $\methodname$ contributes to this goal by offering a schema that aids both annotators and reviewers in assessing dataset quality and utility at a glance.

\section*{Acknowledgments}
We would like to thank Graham Neubig for his valuable insights and constructive discussions, which were instrumental in shaping our rubric design. We also appreciate Blair Yang for helpful conversations regarding the experimental setup.


\bibliography{main}

\appendix
\newpage



\section{Paper Statistics via Automatic Evaluation}

We present paper trends from 2022 to 2024, based on data generated by GPT-4.1 mini using the $\methodname$ framework. The orange, blue, and dark blue bars indicate papers from 2022, 2023, and 2024, respectively.

\subsection{Data Annotations}
\begin{figure}[!ht]
    \centering
    \begin{subfigure}[b]{0.95\textwidth}
        \includegraphics[width=\textwidth]{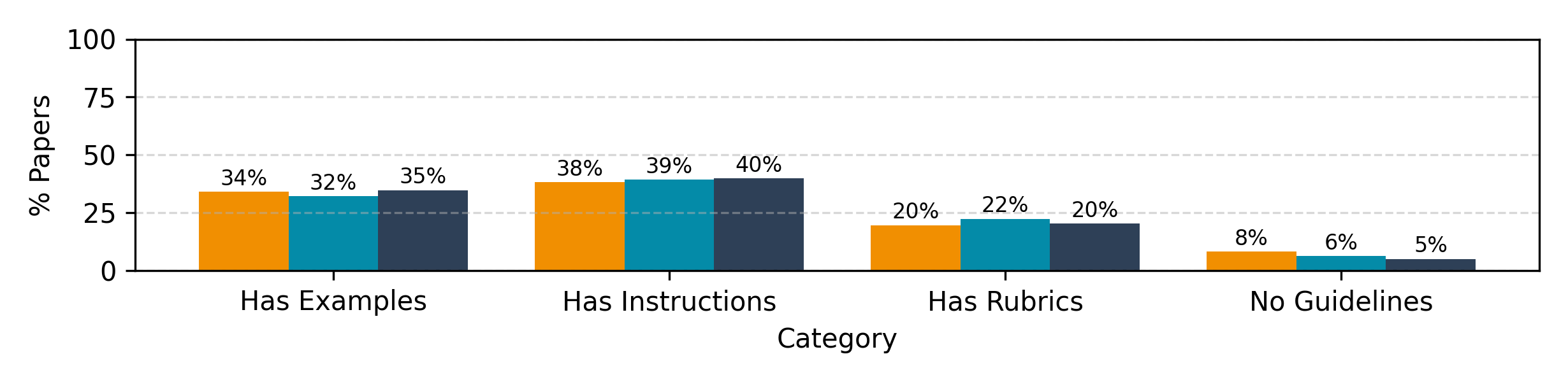}
        \caption{Trends in Data Annotations of Papers at ACL (2022-2024).}
        \label{fig:data-annotation-acl}
    \end{subfigure}
    \begin{subfigure}[b]{0.95\textwidth}
        \includegraphics[width=\textwidth]{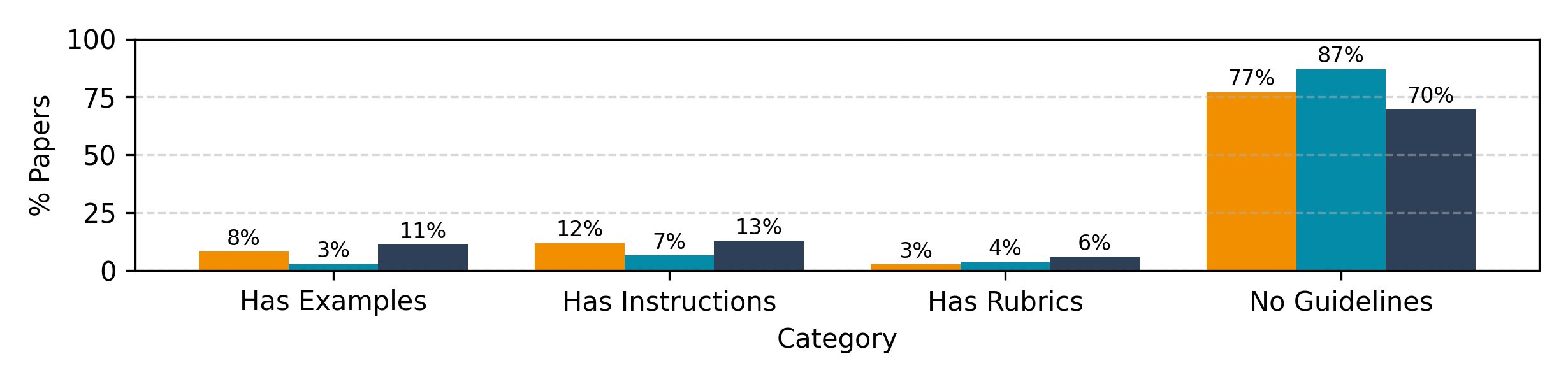}
        \caption{Trends in Data Annotations of Papers at CVPR (2022-2024).}
        \label{fig:data-annotation-cvpr}
    \end{subfigure}
    \begin{subfigure}[b]{0.95\textwidth}
        \includegraphics[width=\textwidth]{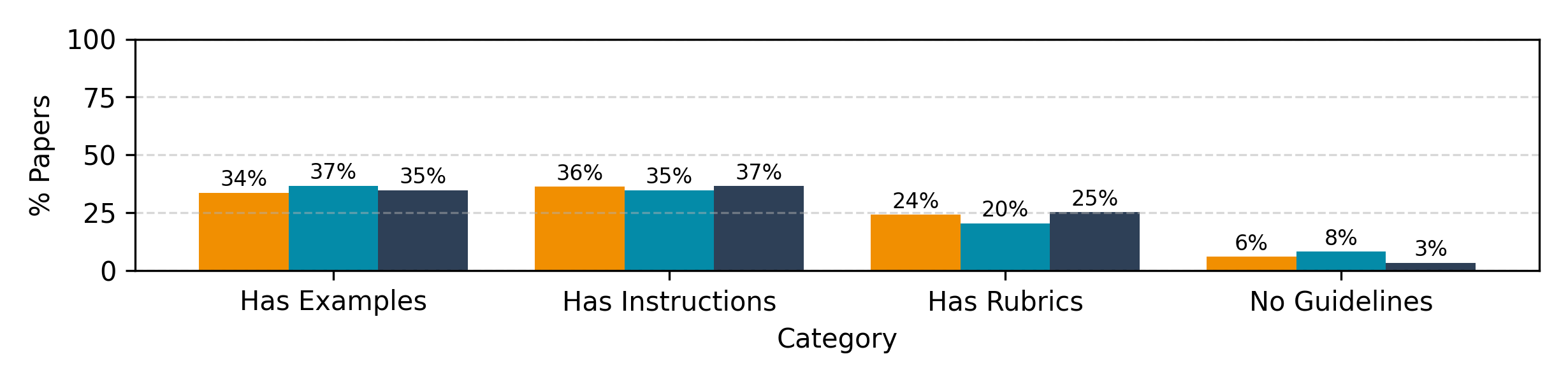}
        \caption{Trends in Data Annotations of Papers at EMNLP (2022-2024).}
        \label{fig:data-annotation-emnlp}
    \end{subfigure}
    \begin{subfigure}[b]{0.95\textwidth}
        \includegraphics[width=\textwidth]{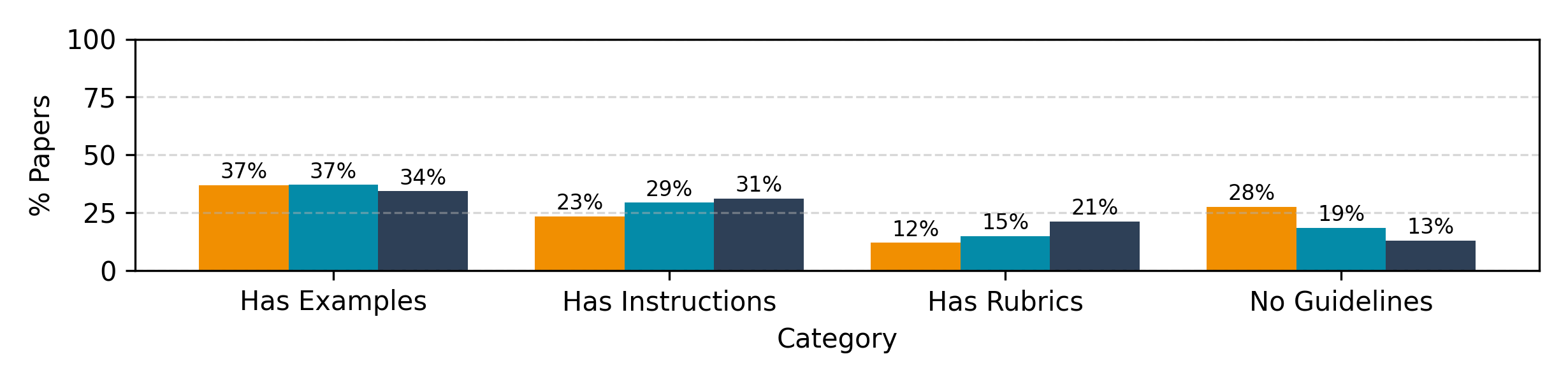}
        \caption{Trends in Data Annotations of Papers at ICLR (2022-2024).}
        \label{fig:data-annotation-iclr}
    \end{subfigure}
    \caption{Paper statistics across conferences on Data Annotations at ACL, CVPR, EMNLP, and ICLR (2022-2024).}
    \label{fig:paper-statistics-data-quality-part1}
\end{figure}

\begin{figure}[!ht]
    \centering
    \begin{subfigure}[b]{0.95\textwidth}
        \includegraphics[width=\textwidth]{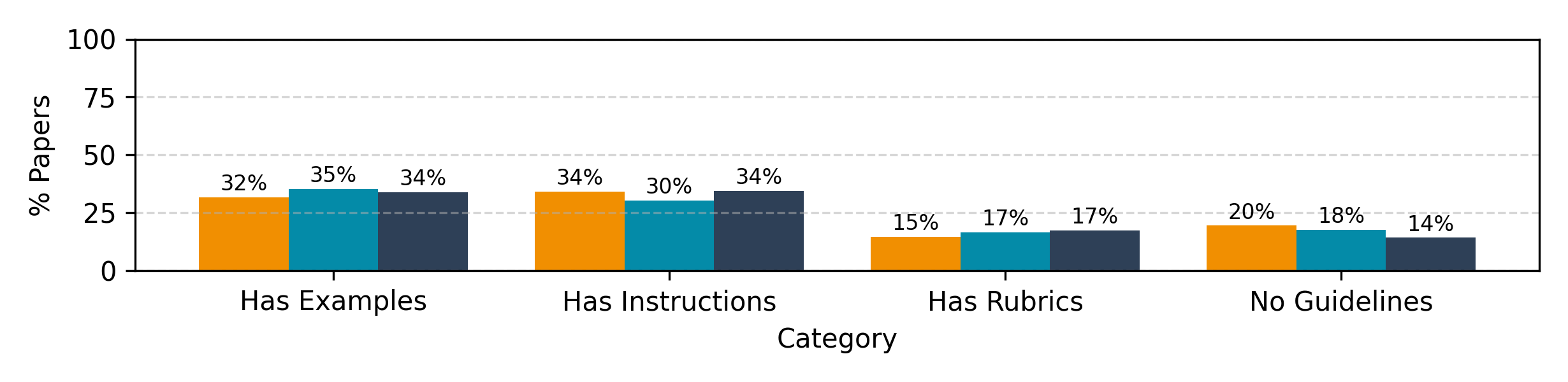}
        \caption{Trends in Data Annotations of Papers at ICML (2022-2024).}
        \label{fig:data-annotation-icml}
    \end{subfigure}
    \begin{subfigure}[b]{0.95\textwidth}
        \includegraphics[width=\textwidth]{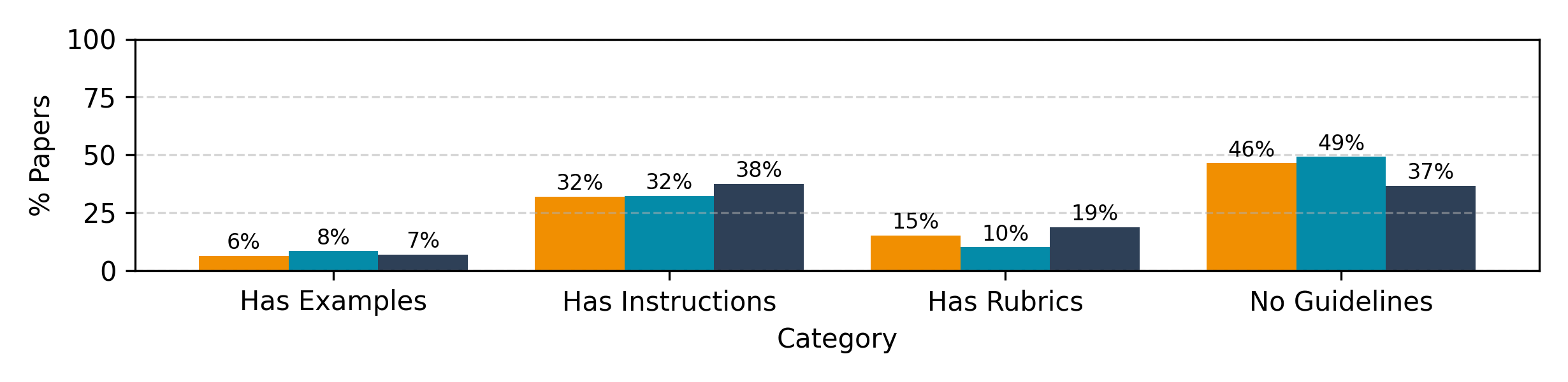}
        \caption{Trends in Data Annotations of Papers at Interspeech (2022-2024).}
        \label{fig:data-annotation-interspeech}
    \end{subfigure}
    \begin{subfigure}[b]{0.95\textwidth}
        \includegraphics[width=\textwidth]{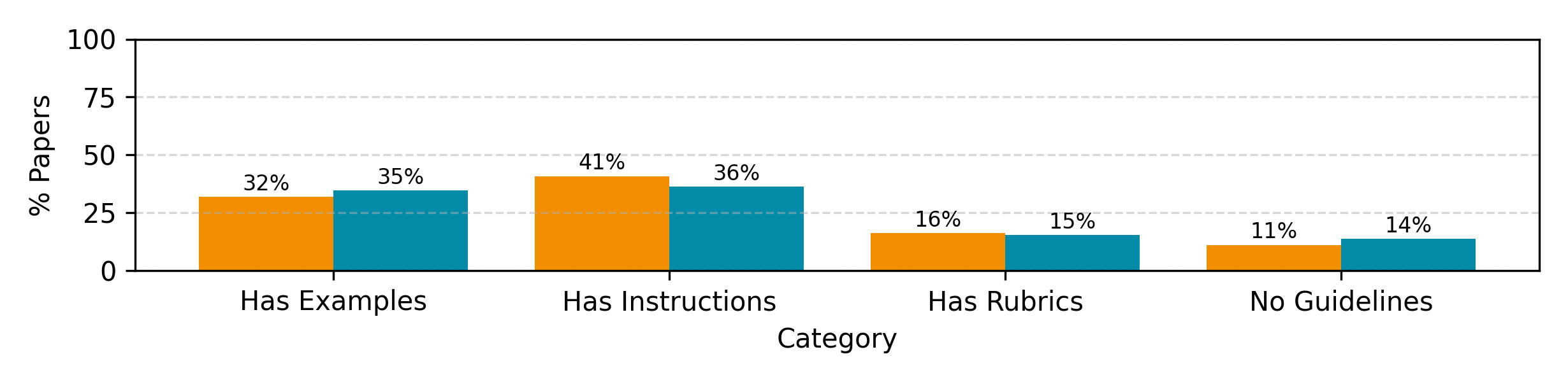}
        \caption{Trends in Data Annotations of Papers at LREC (2022-2024).}
        \label{fig:data-annotation-lrec}
    \end{subfigure}
    \begin{subfigure}[b]{0.95\textwidth}
        \includegraphics[width=\textwidth]{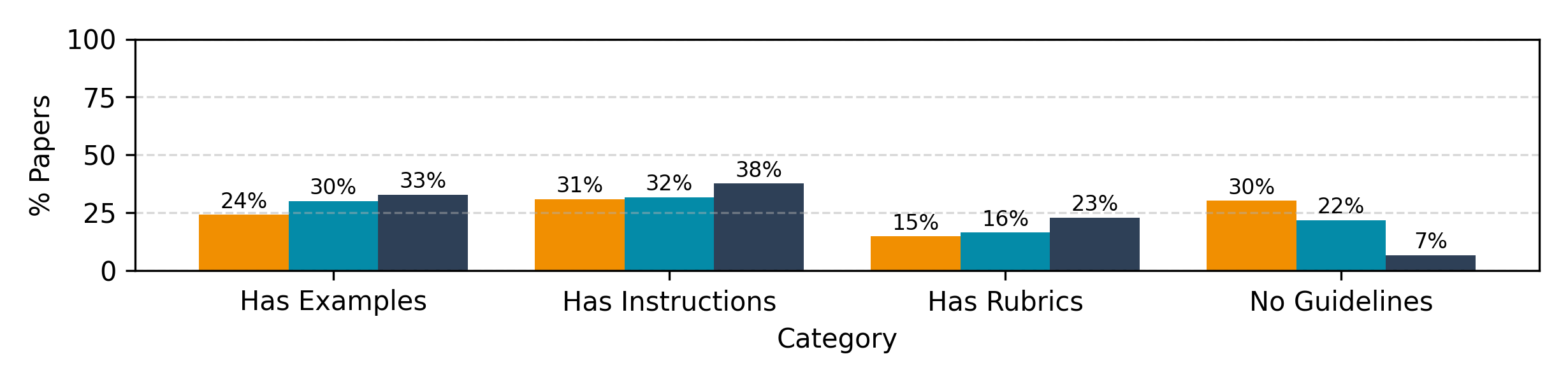}
        \caption{Trends in Data Annotations of Papers at NeurIPS (2022-2024).}
        \label{fig:data-annotation-neurips}
    \end{subfigure}
    \caption{Paper statistics across conferences on Data Annotations at ICML, Interspeech, LREC, and NeurIPS (2022-2024).}
    \label{fig:paper-statistics-data-quality-part2}
\end{figure}

Figures \ref{fig:paper-statistics-data-quality-part1} and \ref{fig:paper-statistics-data-quality-part2} show the paper trends on \textbf{Data Annotations} at ACL, CVPR, EMNLP, ICLR, ICML, Interspeech, LREC, and NeurIPS.

\subsection{Data Novelty}
\begin{figure}[!ht]
    \centering
    \begin{subfigure}[b]{\textwidth}
        \includegraphics[width=\textwidth]{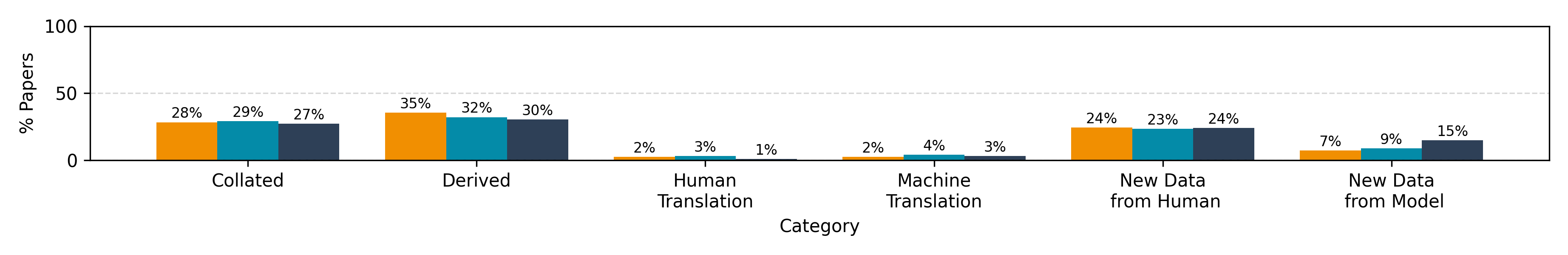}
        \caption{Trends in Data Novelty of Papers at ACL (2022-2024).}
        \label{fig:data-novelty-acl}
    \end{subfigure}
    \begin{subfigure}[b]{\textwidth}
        \includegraphics[width=\textwidth]{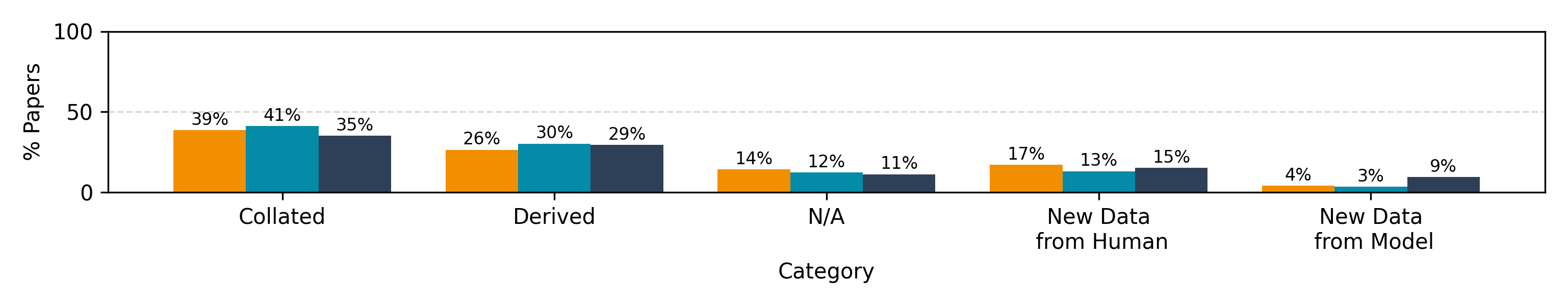}
        \caption{Trends in Data Novelty of Papers at CVPR (2022-2024).}
        \label{fig:data-novelty-cvpr}
    \end{subfigure}
    \begin{subfigure}[b]{\textwidth}
        \includegraphics[width=\textwidth]{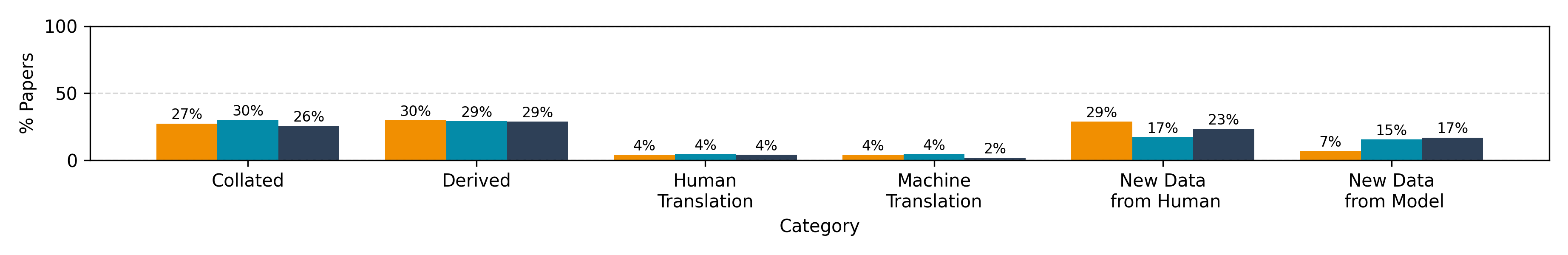}
        \caption{Trends in Data Novelty of Papers at EMNLP (2022-2024).}
        \label{fig:data-novelty-emnlp}
    \end{subfigure}
    \begin{subfigure}[b]{\textwidth}
        \includegraphics[width=\textwidth]{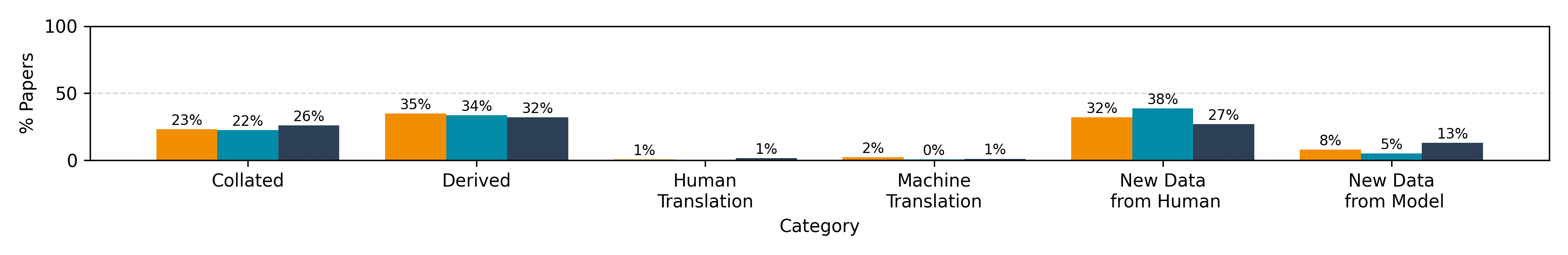}
        \caption{Trends in Data Novelty of Papers at ICLR (2022-2024).}
        \label{fig:data-novelty-iclr}
    \end{subfigure}
    \caption{Paper statistics across conferences on Data Novelty at ACL, CVPR, EMNLP, and ICLR (2022-2024).}
    \label{fig:paper-statistics-data-novelty-part1}
\end{figure}

\begin{figure}[!ht]
    \centering
    \begin{subfigure}[b]{\textwidth}
        \includegraphics[width=\textwidth]{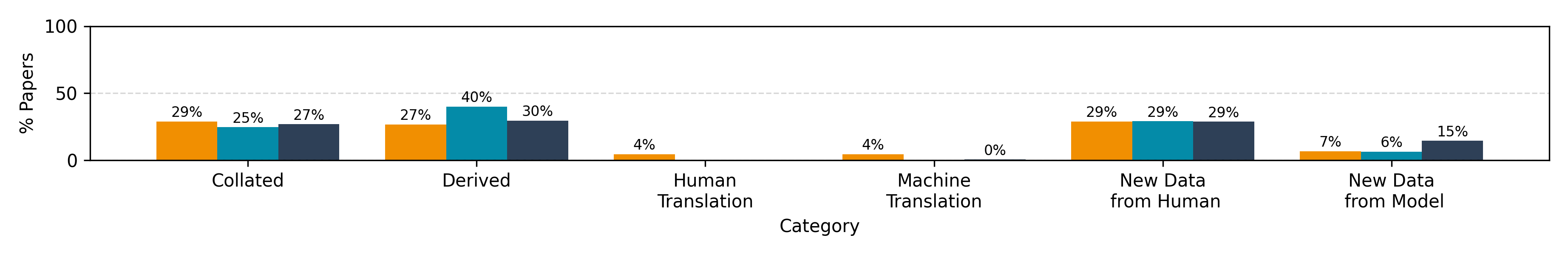}
        \caption{Trends in Data Novelty of Papers at ICML (2022-2024).}
        \label{fig:data-novelty-icml}
    \end{subfigure}
    \begin{subfigure}[b]{\textwidth}
        \includegraphics[width=\textwidth]{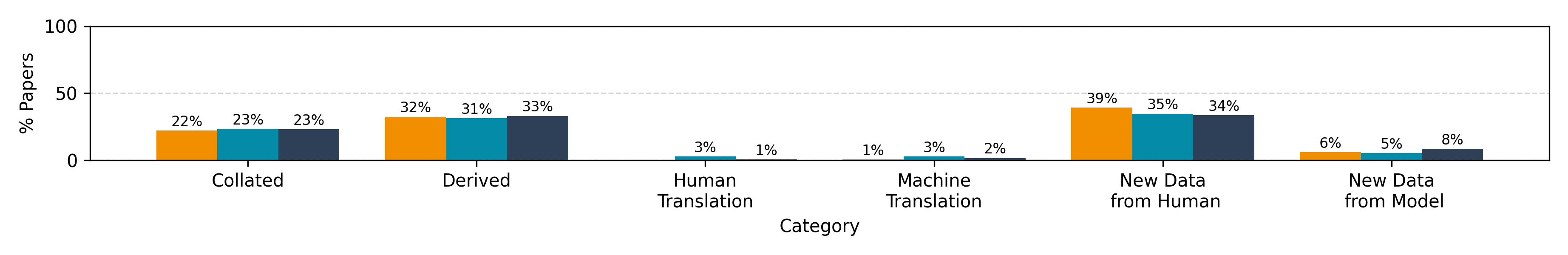}
        \caption{Trends in Data Novelty of Papers at Interspeech (2022-2024).}
        \label{fig:data-novelty-interspeech}
    \end{subfigure}
    \begin{subfigure}[b]{\textwidth}
        \includegraphics[width=\textwidth]{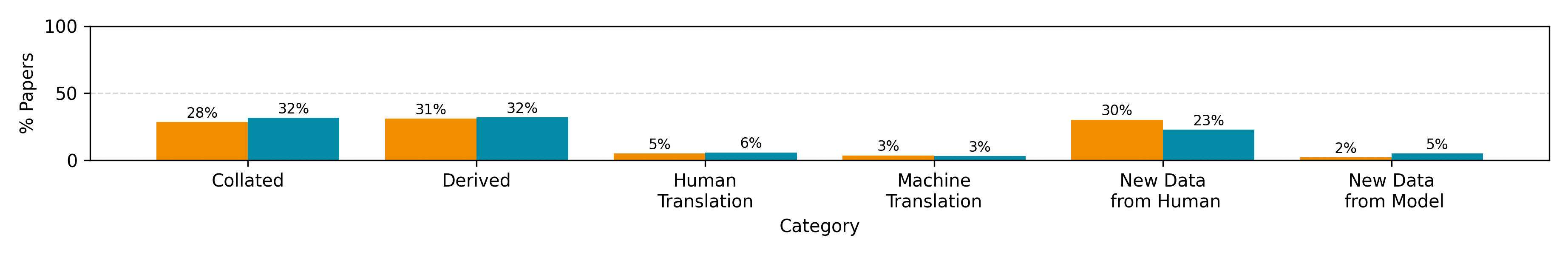}
        \caption{Trends in Data Novelty of Papers at LREC (2022-2024).}
        \label{fig:data-novelty-lrec}
    \end{subfigure}
    \begin{subfigure}[b]{\textwidth}
        \includegraphics[width=\textwidth]{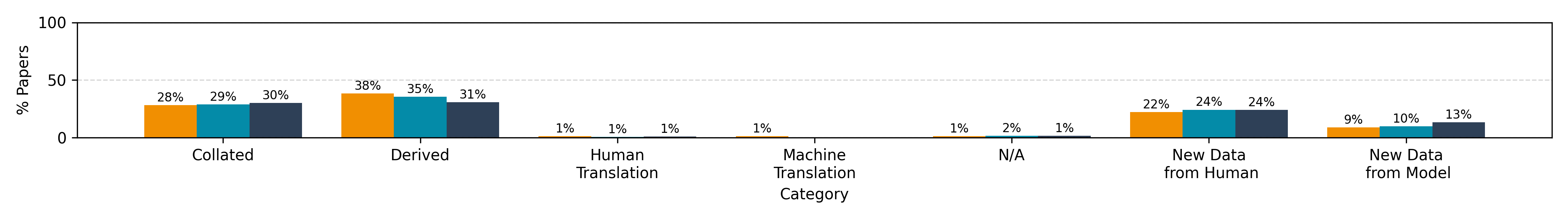}
        \caption{Trends in Data Novelty of Papers at NeurIPS (2022-2024).}
        \label{fig:data-novelty-neurips}
    \end{subfigure}
    \caption{Paper statistics across conferences on Data Novelty at ICML, Interspeech, LREC, and NeurIPS (2022-2024).}
    \label{fig:paper-statistics-data-novelty-part2}
\end{figure}

Figures \ref{fig:paper-statistics-data-novelty-part1} and \ref{fig:paper-statistics-data-novelty-part2} show the paper trends on \textbf{Data Novelty} at ACL, CVPR, EMNLP, ICLR, ICML, Interspeech, LREC, and NeurIPS.

\subsection{Languages}
\begin{figure}[!ht]
    \centering
    \begin{subfigure}[b]{\textwidth}
        \includegraphics[width=\textwidth]{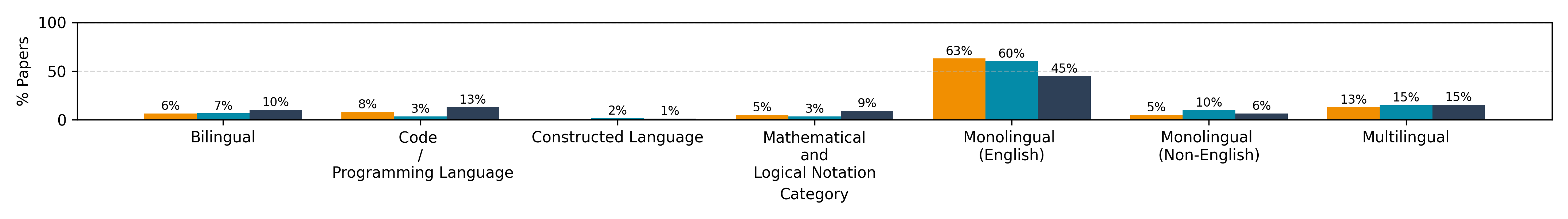}
        \caption{Trends in Languages of Papers at ACL.}
        \label{fig:languages-acl}
    \end{subfigure}
    \begin{subfigure}[b]{\textwidth}
        \includegraphics[width=\textwidth]{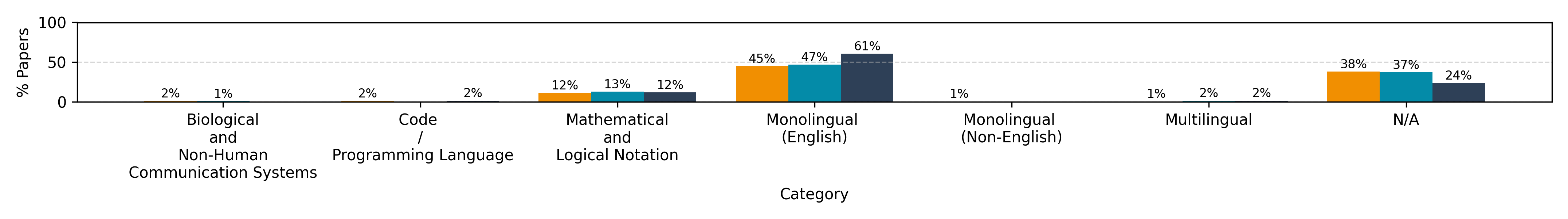}
        \caption{Trends in Languages of Papers at CVPR.}
        \label{fig:languages-cvpr}
    \end{subfigure}
    \begin{subfigure}[b]{\textwidth}
        \includegraphics[width=\textwidth]{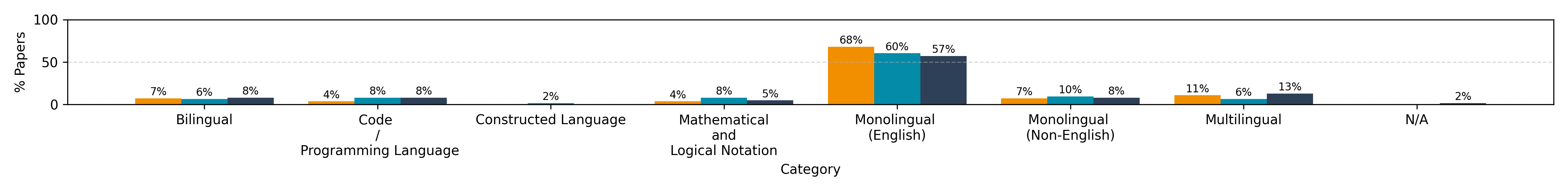}
        \caption{Trends in Languages of Papers at EMNLP.}
        \label{fig:languages-emnlp}
    \end{subfigure}
    \begin{subfigure}[b]{\textwidth}
        \includegraphics[width=\textwidth]{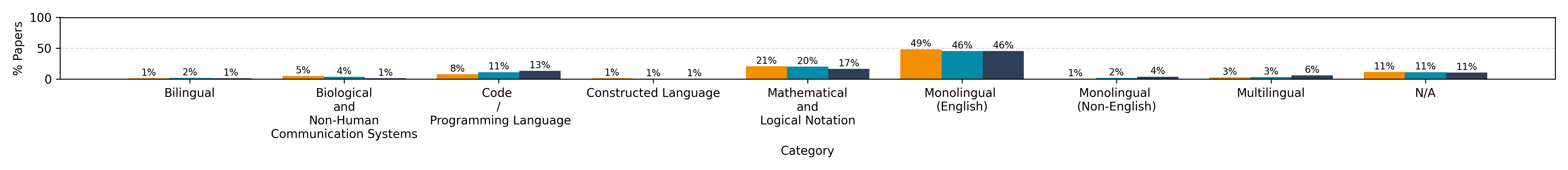}
        \caption{Trends in Languages of Papers at ICLR.}
        \label{fig:languages-iclr}
    \end{subfigure}
    \caption{Paper statistics across conferences on Languages at ACL, CVPR, EMNLP, and ICLR (2022-2024).}
    \label{fig:paper-statistics-languages-part1}
\end{figure}

\begin{figure}[!ht]
    \centering
    \begin{subfigure}[b]{\textwidth}
        \includegraphics[width=\textwidth]{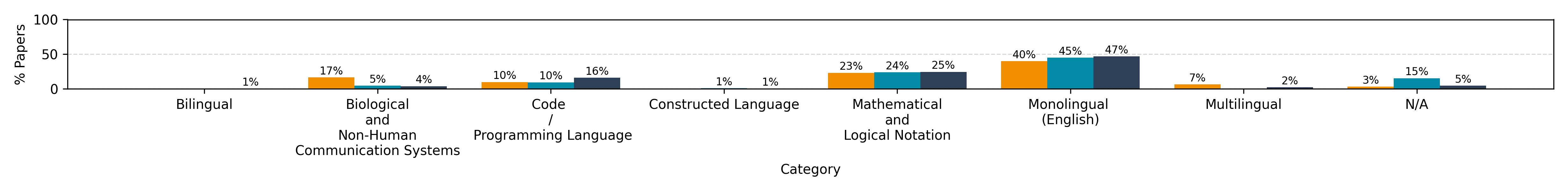}
        \caption{Trends in Languages of Papers at ICML.}
        \label{fig:languages-icml}
    \end{subfigure}
    \begin{subfigure}[b]{\textwidth}
        \includegraphics[width=\textwidth]{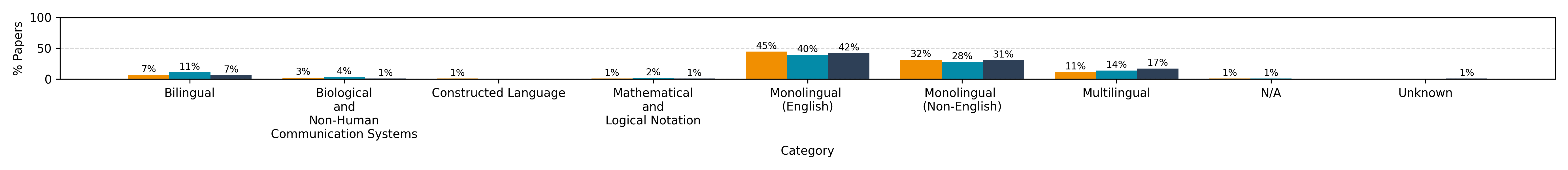}
        \caption{Trends in Languages of Papers at Interspeech.}
        \label{fig:languages-interspeech}
    \end{subfigure}
    \begin{subfigure}[b]{\textwidth}
        \includegraphics[width=\textwidth]{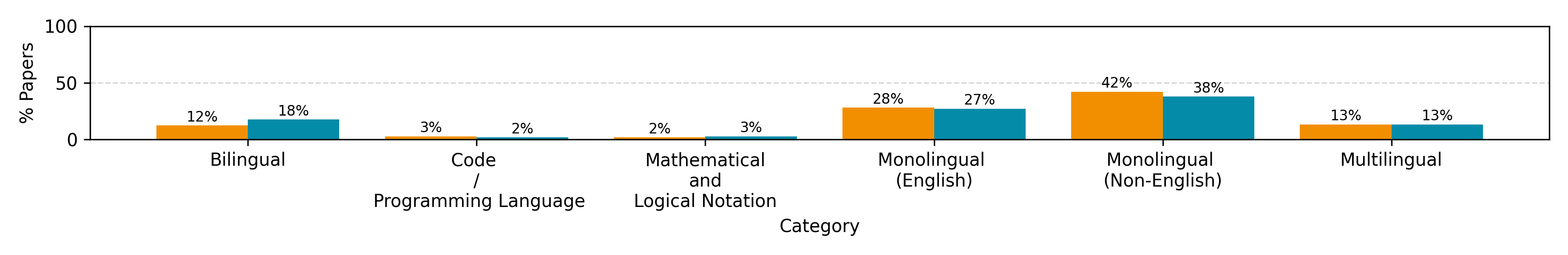}
        \caption{Trends in Languages of Papers at LREC.}
        \label{fig:languages-lrec}
    \end{subfigure}
    \begin{subfigure}[b]{\textwidth}
        \includegraphics[width=\textwidth]{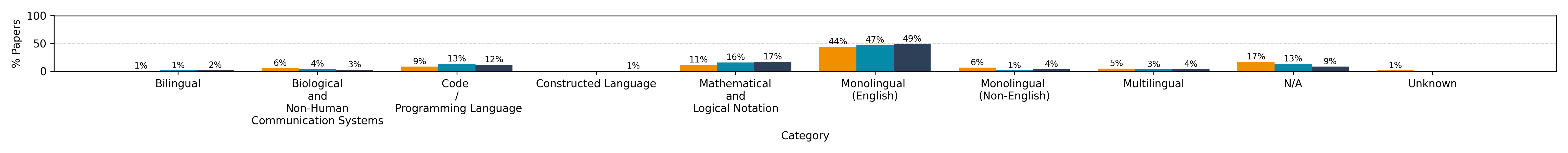}
        \caption{Trends in Languages of Papers at NeurIPS.}
        \label{fig:languages-neurips}
    \end{subfigure}
    \caption{Paper statistics across conferences on Languages at ICML, Interspeech, LREC, and NeurIPS (2022-2024).}
    \label{fig:paper-statistics-languages-part2}
\end{figure}

Figures \ref{fig:paper-statistics-languages-part1} and \ref{fig:paper-statistics-languages-part2} show the paper trends on \textbf{Languages} at ACL, CVPR, EMNLP, ICLR, ICML, Interspeech, LREC, and NeurIPS.

\subsection{Quality Assurance}
\begin{figure}[!ht]
    \centering
    \begin{subfigure}[b]{\textwidth}
        \includegraphics[width=\textwidth]{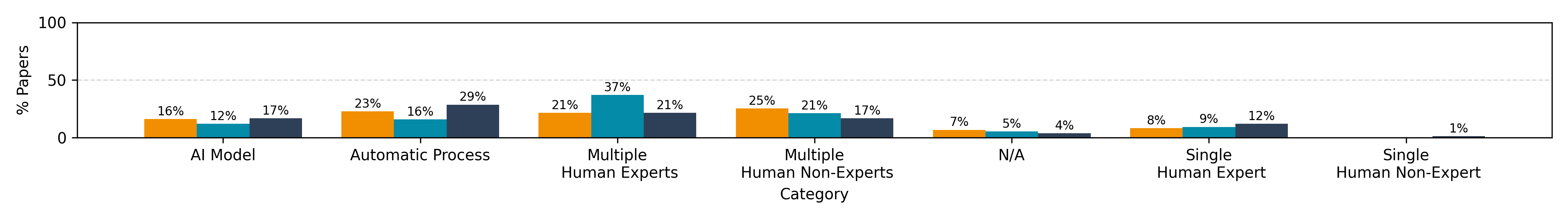}
        \caption{Trends in Quality Assurance of Papers at ACL.}
        \label{fig:qa-acl}
    \end{subfigure}
    \begin{subfigure}[b]{\textwidth}
        \includegraphics[width=\textwidth]{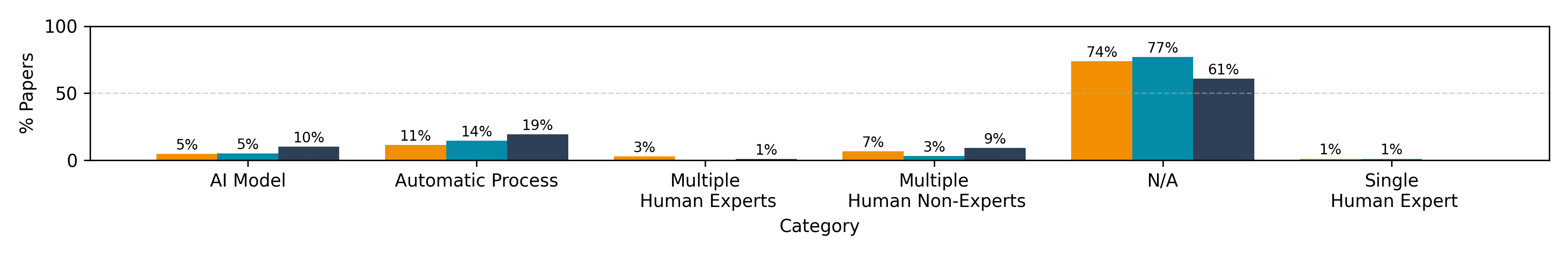}
        \caption{Trends in Quality Assurance of Papers at CVPR.}
        \label{fig:qa-cvpr}
    \end{subfigure}
    \begin{subfigure}[b]{\textwidth}
        \includegraphics[width=\textwidth]{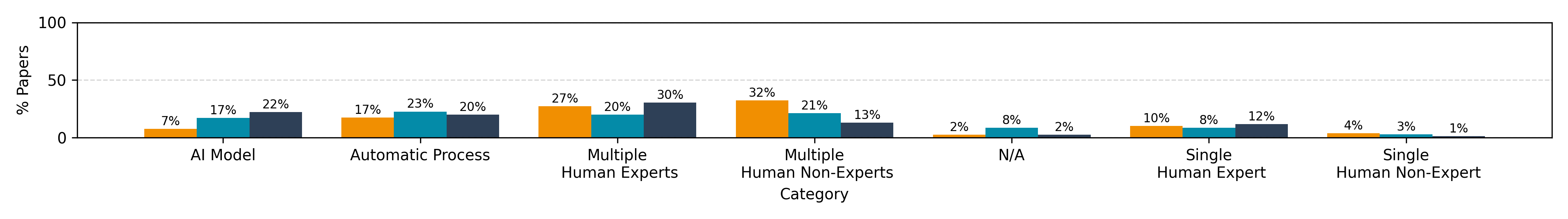}
        \caption{Trends in Quality Assurance of Papers at EMNLP.}
        \label{fig:qa-emnlp}
    \end{subfigure}
    \begin{subfigure}[b]{\textwidth}
        \includegraphics[width=\textwidth]{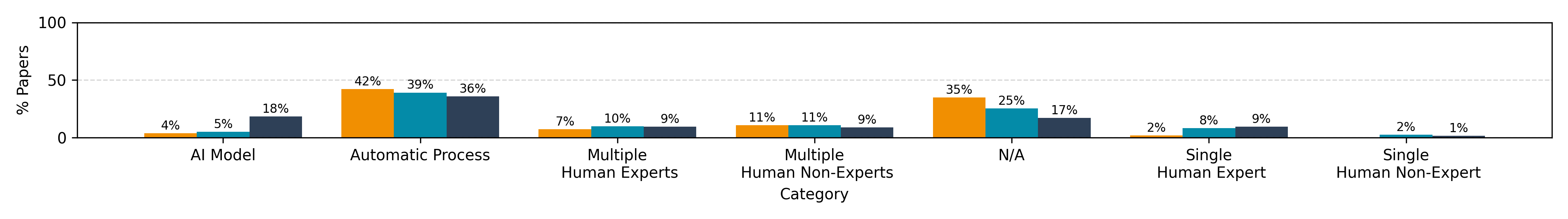}
        \caption{Trends in Quality Assurance of Papers at ICLR.}
        \label{fig:qa-iclr}
    \end{subfigure}
    \caption{Paper statistics across conferences on Quality Assurance at ACL, CVPR, EMNLP, and ICLR (2022-2024).}
    \label{fig:paper-statistics-qa-part1}
\end{figure}

\begin{figure}[!ht]
    \centering
    \begin{subfigure}[b]{\textwidth}
        \includegraphics[width=\textwidth]{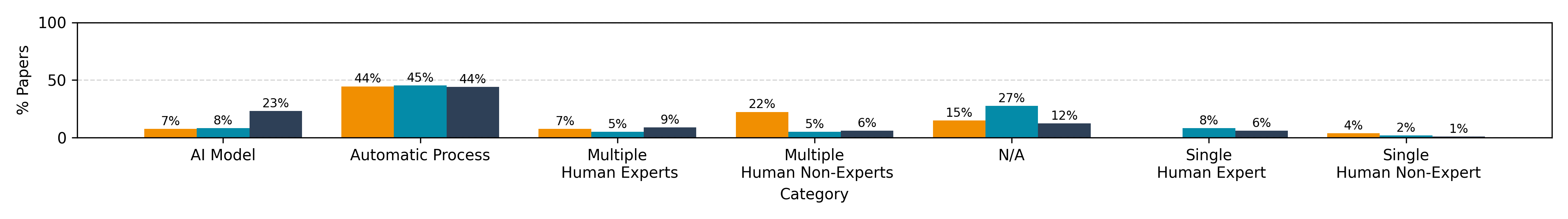}
        \caption{Trends in Quality Assurance of Papers at ICML.}
        \label{fig:qa-icml}
    \end{subfigure}
    \begin{subfigure}[b]{\textwidth}
        \includegraphics[width=\textwidth]{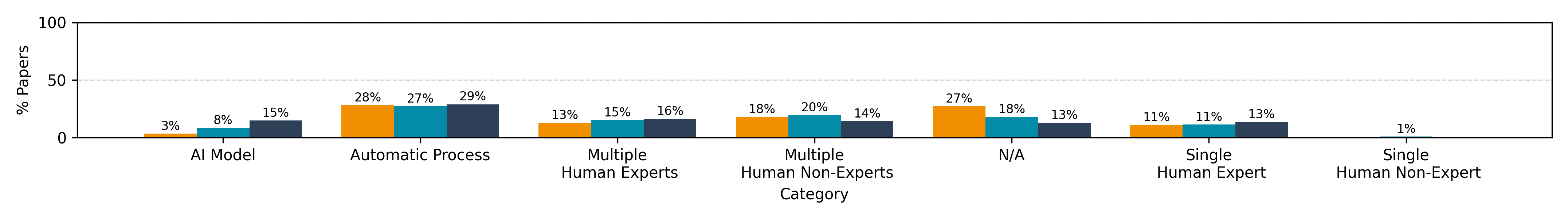}
        \caption{Trends in Quality Assurance of Papers at Interspeech.}
        \label{fig:qa-interspeech}
    \end{subfigure}
    \begin{subfigure}[b]{\textwidth}
        \includegraphics[width=\textwidth]{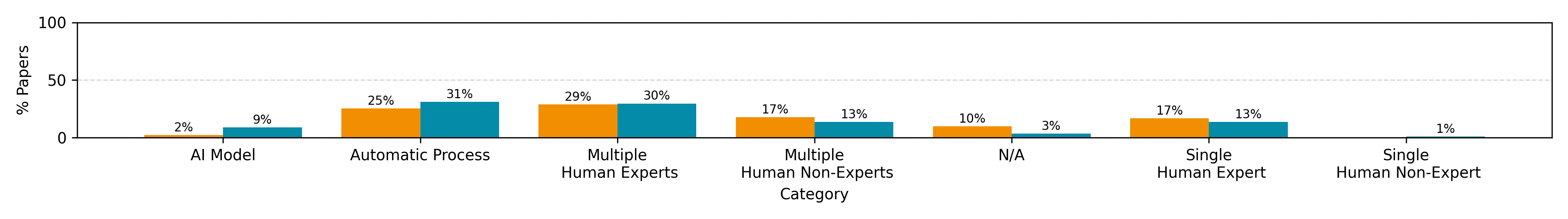}
        \caption{Trends in Quality Assurance of Papers at LREC.}
        \label{fig:qa-lrec}
    \end{subfigure}
    \begin{subfigure}[b]{\textwidth}
        \includegraphics[width=\textwidth]{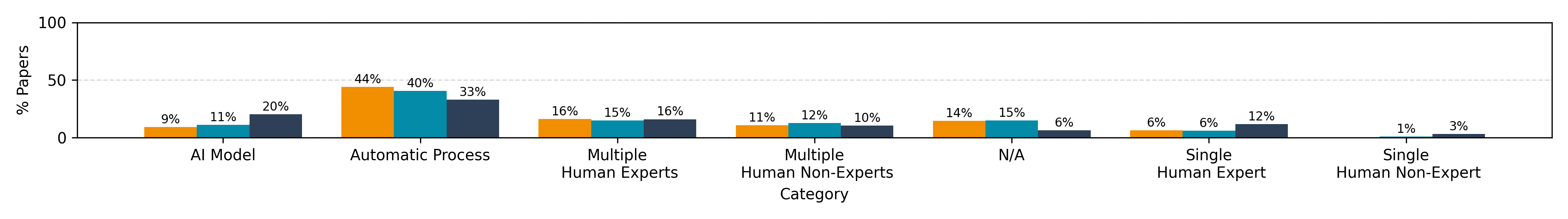}
        \caption{Trends in Quality Assurance of Papers at NeurIPS.}
        \label{fig:qa-neurips}
    \end{subfigure}
    \caption{Paper statistics across conferences on Quality Assurance at ICML, Interspeech, LREC, and NeurIPS (2022-2024).}
    \label{fig:paper-statistics-qa-part2}
\end{figure}

Figures \ref{fig:paper-statistics-qa-part1} and \ref{fig:paper-statistics-qa-part2} show the paper trends on \textbf{Quality Assurance} at ACL, CVPR, EMNLP, ICLR, ICML, Interspeech, LREC, and NeurIPS.

\subsection{Reproducibility: Data Creation and Code}
\begin{figure}[!ht]
    \centering
    \begin{subfigure}[b]{0.45\textwidth}
        \includegraphics[width=\textwidth]{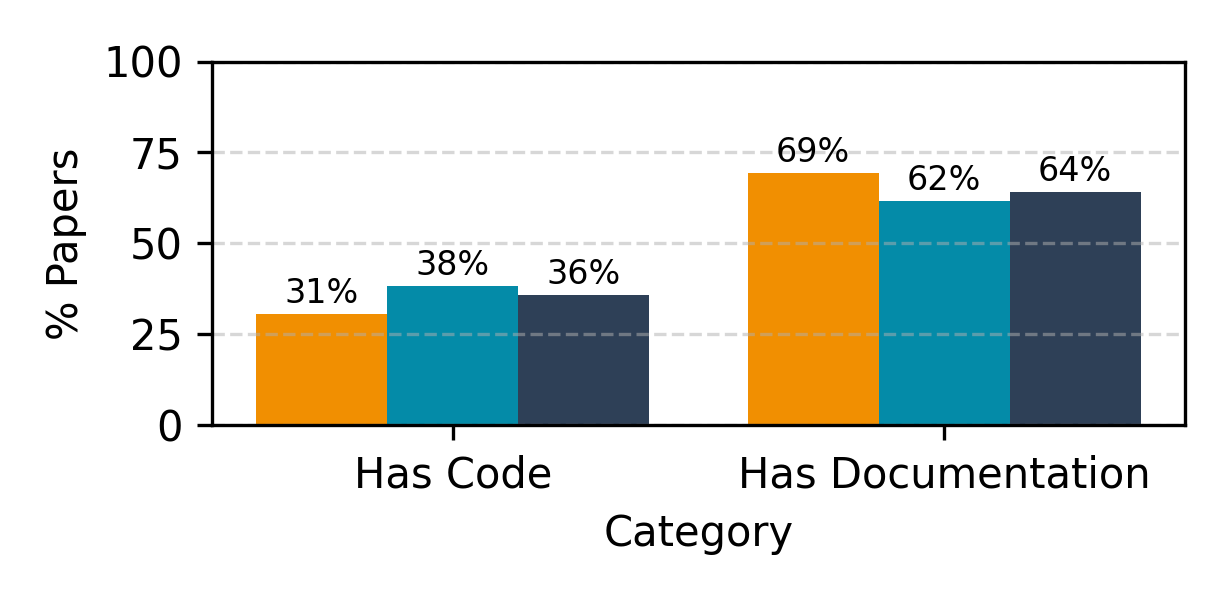}
        \caption{Trends in Data Creation and Code of Papers at ACL.}
        \label{fig:reproducibility-acl}
    \end{subfigure}
    \hfill
    \begin{subfigure}[b]{0.45\textwidth}
        \includegraphics[width=\textwidth]{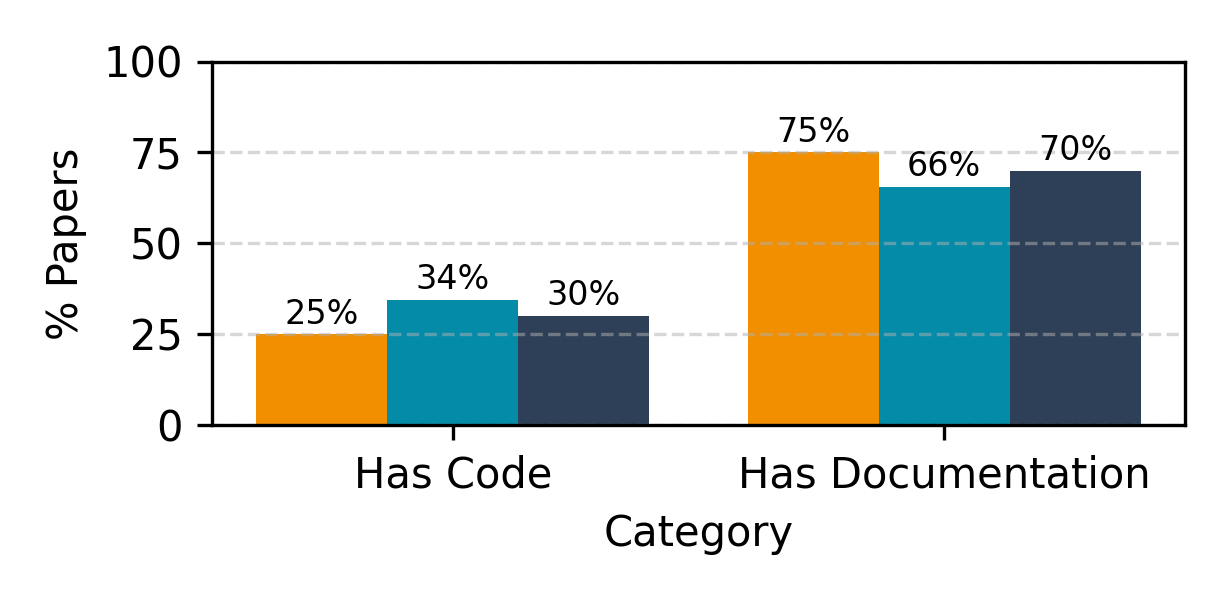}
        \caption{Trends in Data Creation and Code of Papers at CVPR.}
        \label{fig:reproducibility-cvpr}
    \end{subfigure}
    \hfill
    \begin{subfigure}[b]{0.45\textwidth}
        \includegraphics[width=\textwidth]{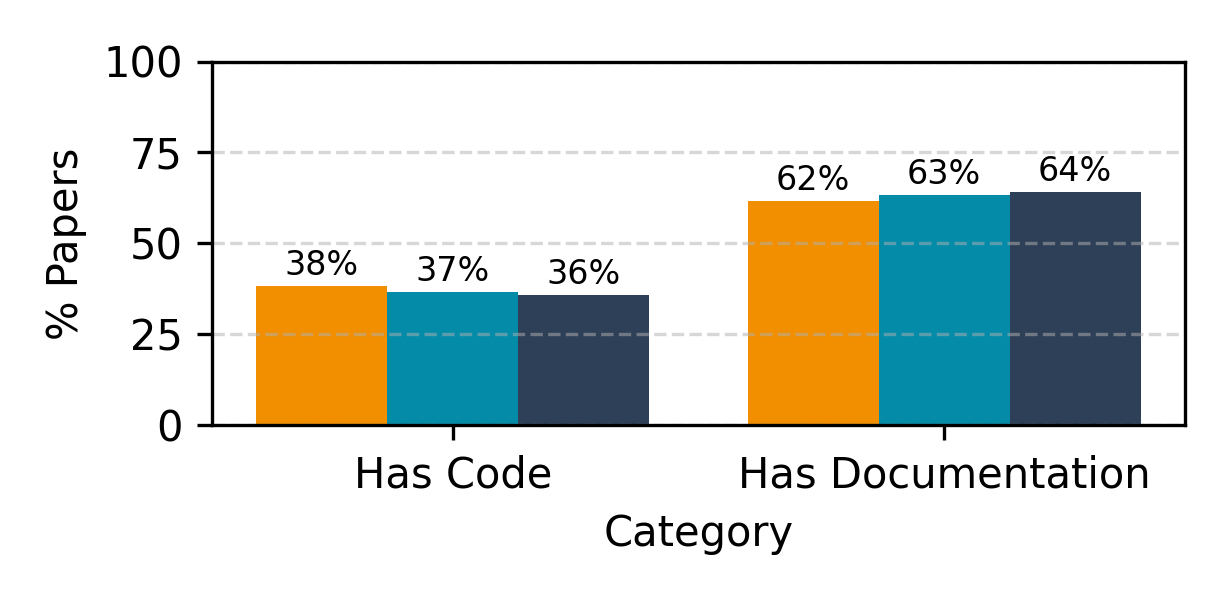}
        \caption{Trends in Data Creation and Code of Papers at EMNLP.}
        \label{fig:reproducibility-emnlp}
    \end{subfigure}
    \hfill
    \begin{subfigure}[b]{0.45\textwidth}
        \includegraphics[width=\textwidth]{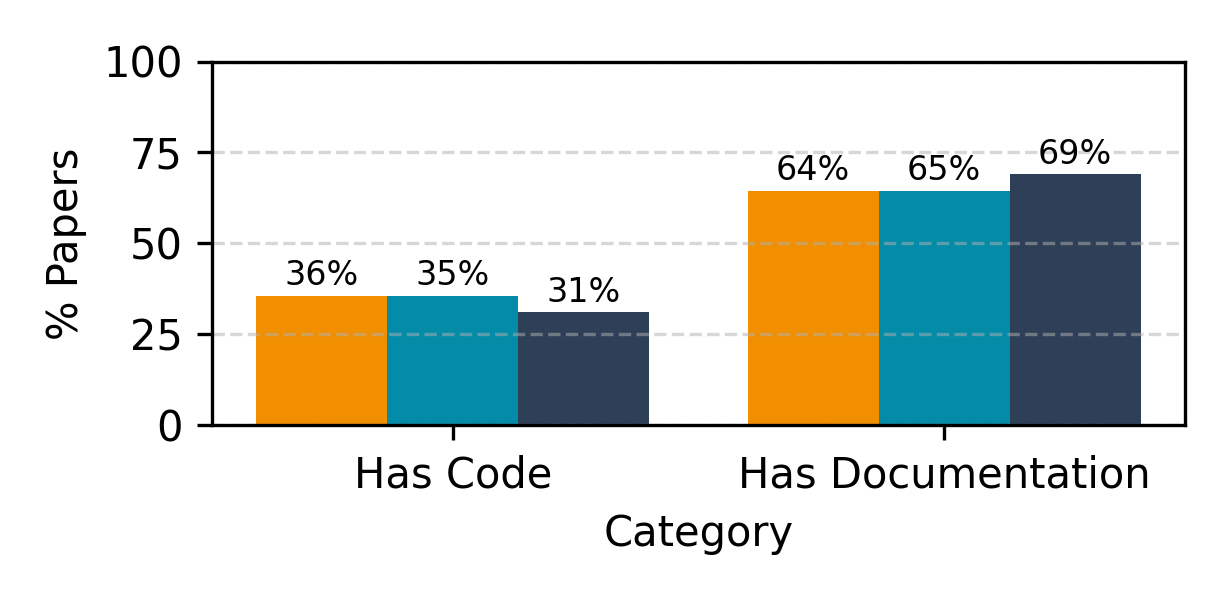}
        \caption{Trends in Data Creation and Code of Papers at ICLR.}
        \label{fig:reproducibility-iclr}
    \end{subfigure}
    \caption{Paper statistics across conferences on Data Creation and Code at ACL, CVPR, EMNLP, and ICLR (2022-2024).}
    \label{fig:paper-statistics-reproducibility-part1}
\end{figure}

\begin{figure}[!ht]
    \centering
    \begin{subfigure}[b]{0.45\textwidth}
        \includegraphics[width=\textwidth]{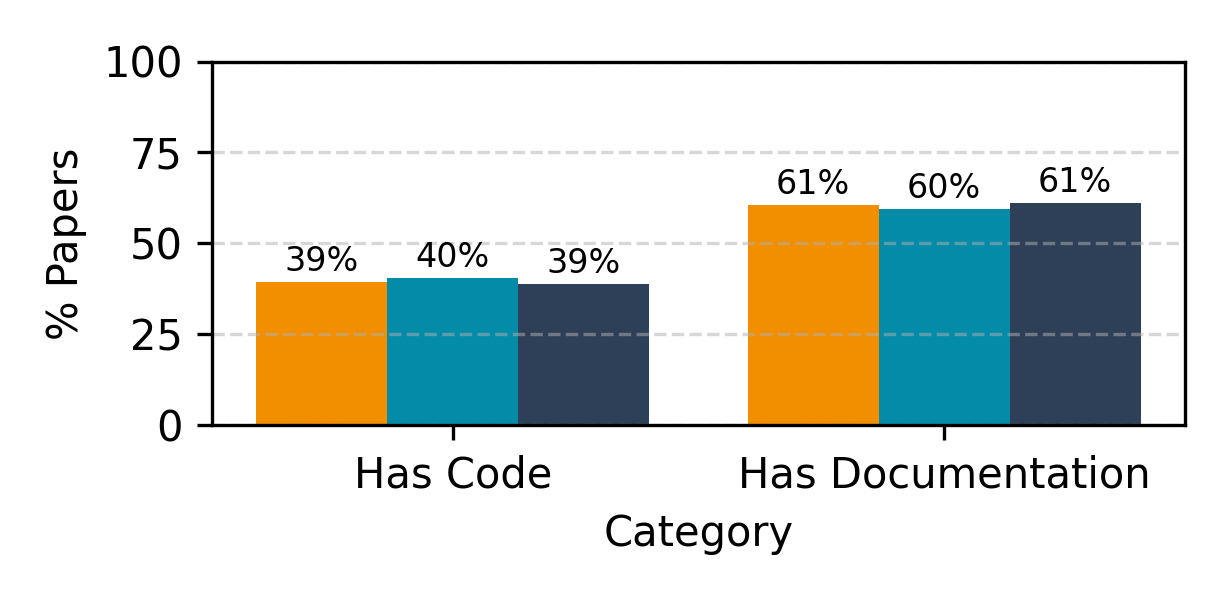}
        \caption{Trends in Data Creation and Code of Papers at ICML.}
        \label{fig:reproducibility-icml}
    \end{subfigure}
    \hfill
    \begin{subfigure}[b]{0.45\textwidth}
        \includegraphics[width=\textwidth]{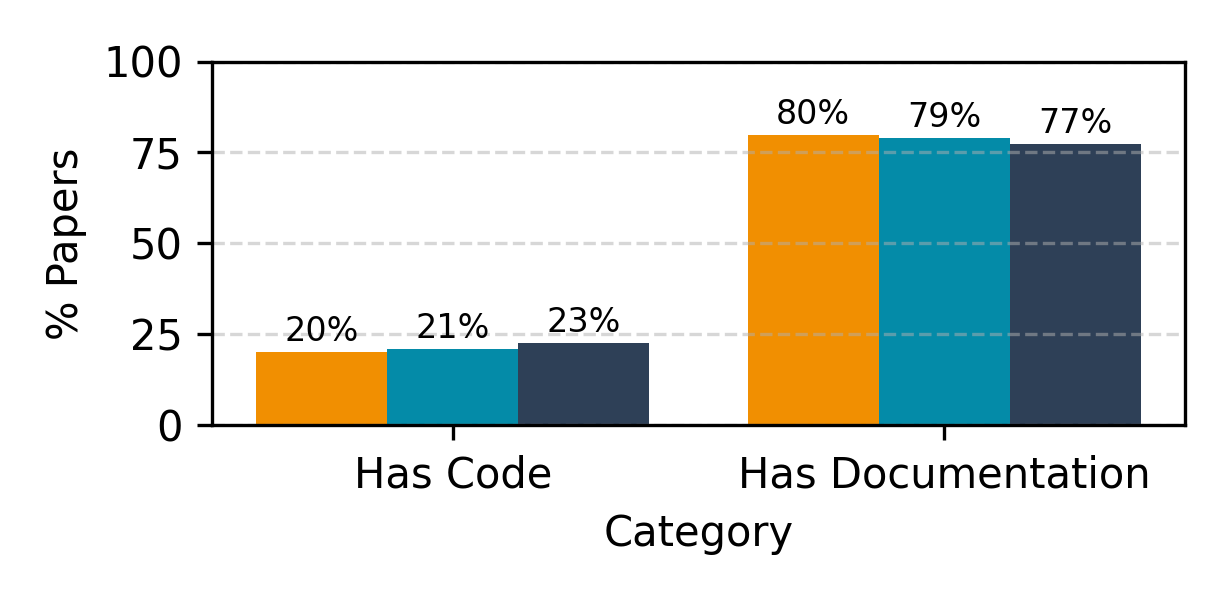}
        \caption{Trends in Data Creation and Code of Papers at Interspeech.}
        \label{fig:reproducibility-interspeech}
    \end{subfigure}
    \hfill
    \begin{subfigure}[b]{0.45\textwidth}
        \includegraphics[width=\textwidth]{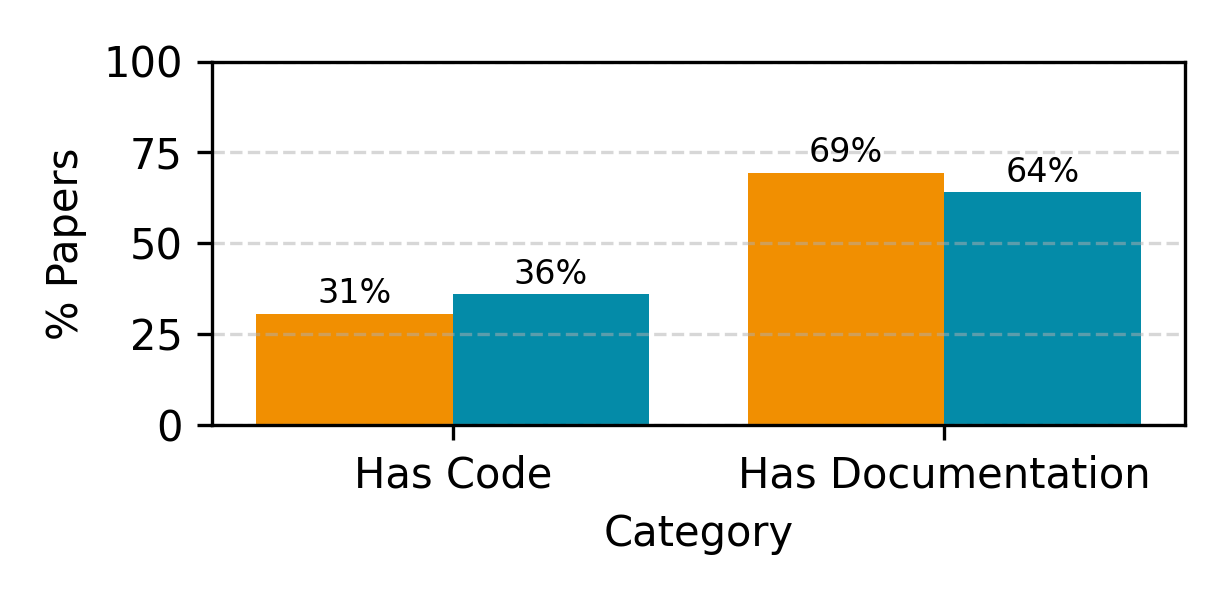}
        \caption{Trends in Data Creation and Code of Papers at LREC.}
        \label{fig:reproducibility-lrec}
    \end{subfigure}
    \hfill
    \begin{subfigure}[b]{0.45\textwidth}
        \includegraphics[width=\textwidth]{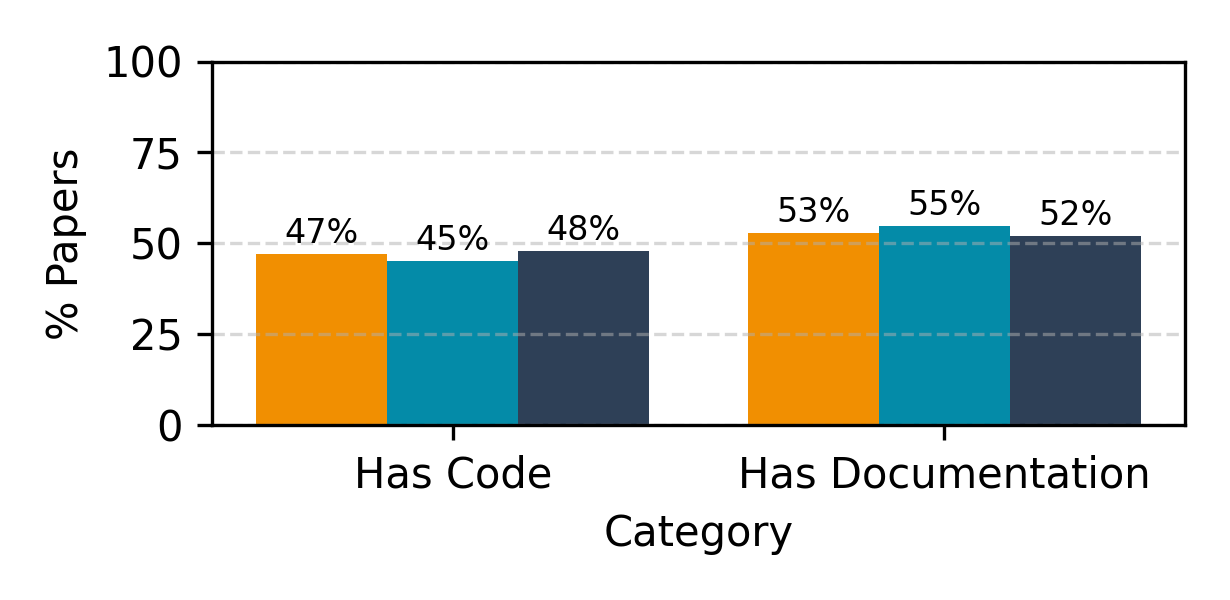}
        \caption{Trends in Data Creation and Code of Papers at NeurIPS.}
        \label{fig:reproducibility-neurips}
    \end{subfigure}
    \caption{Paper statistics across conferences on Data Creation and Code at ICML, Interspeech, LREC, and NeurIPS (2022-2024).}
    \label{fig:paper-statistics-reproducibility-part2}
\end{figure}

Figures \ref{fig:paper-statistics-reproducibility-part1} and \ref{fig:paper-statistics-reproducibility-part2} show the paper trends on \textbf{Data Creation} and \textbf{Code} criterion at ACL, CVPR, EMNLP, ICLR, ICML, Interspeech, LREC, and NeurIPS.

\subsection{Task Utility}
\begin{figure}[!ht]
    \centering
    \begin{subfigure}[b]{\textwidth}
        \includegraphics[width=\textwidth]{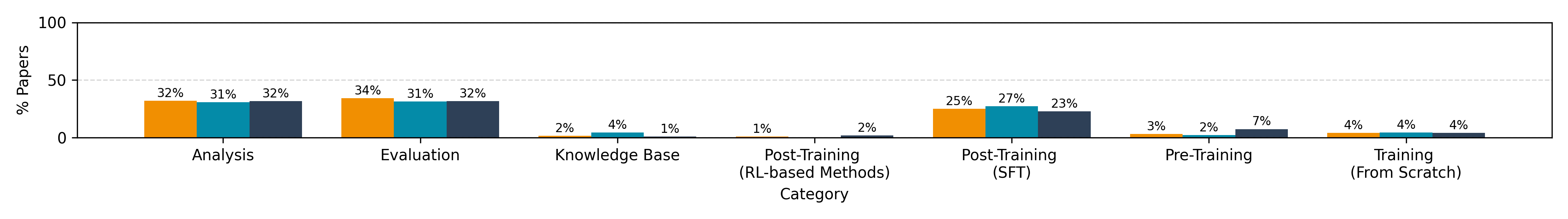}
        \caption{Trends in Task Utility of Papers at ACL.}
        \label{fig:task-utility-acl}
    \end{subfigure}
    \begin{subfigure}[b]{\textwidth}
        \includegraphics[width=\textwidth]{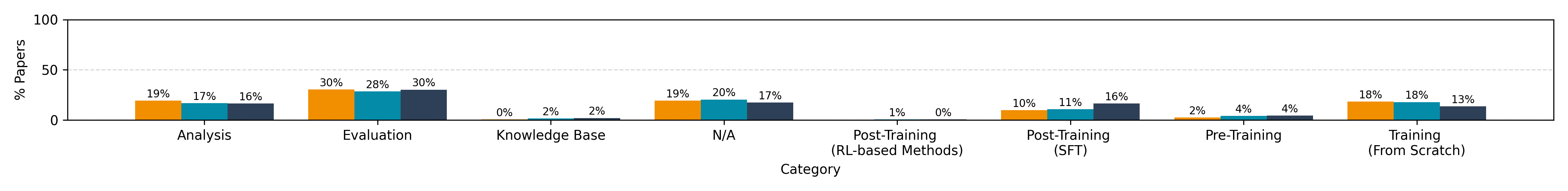}
        \caption{Trends in Task Utility of Papers at CVPR.}
        \label{fig:task-utility-cvpr}
    \end{subfigure}
    \begin{subfigure}[b]{\textwidth}
        \includegraphics[width=\textwidth]{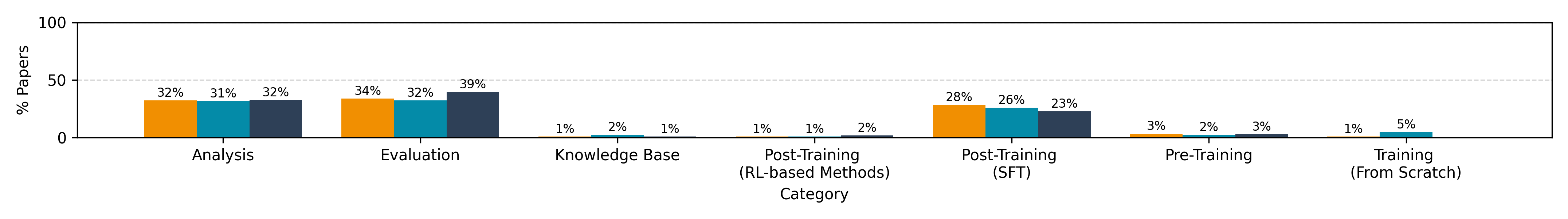}
        \caption{Trends in Task Utility of Papers at EMNLP.}
        \label{fig:task-utility-emnlp}
    \end{subfigure}
    \begin{subfigure}[b]{\textwidth}
        \includegraphics[width=\textwidth]{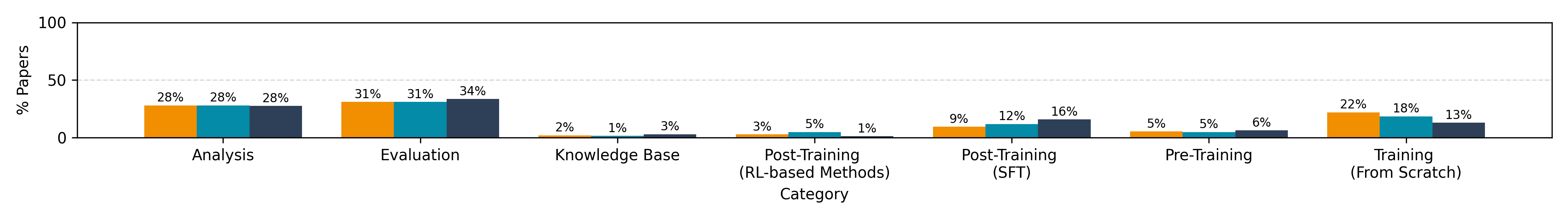}
        \caption{Trends in Task Utility of Papers at ICLR.}
        \label{fig:task-utility-iclr}
    \end{subfigure}
    \caption{Paper statistics across conferences on Task Utility at ACL, CVPR, EMNLP, and ICLR (2022-2024).}
    \label{fig:paper-statistics-task-utility-part1}
\end{figure}

\begin{figure}[!ht]
    \centering
    \begin{subfigure}[b]{\textwidth}
        \includegraphics[width=\textwidth]{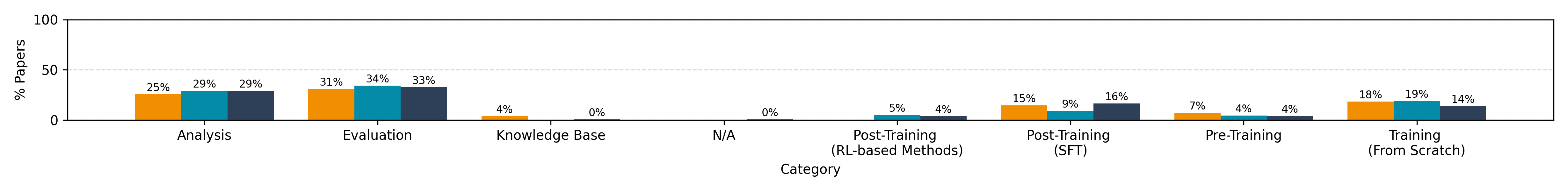}
        \caption{Trends in Task Utility of Papers at ICML.}
        \label{fig:task-utility-icml}
    \end{subfigure}
    \begin{subfigure}[b]{\textwidth}
        \includegraphics[width=\textwidth]{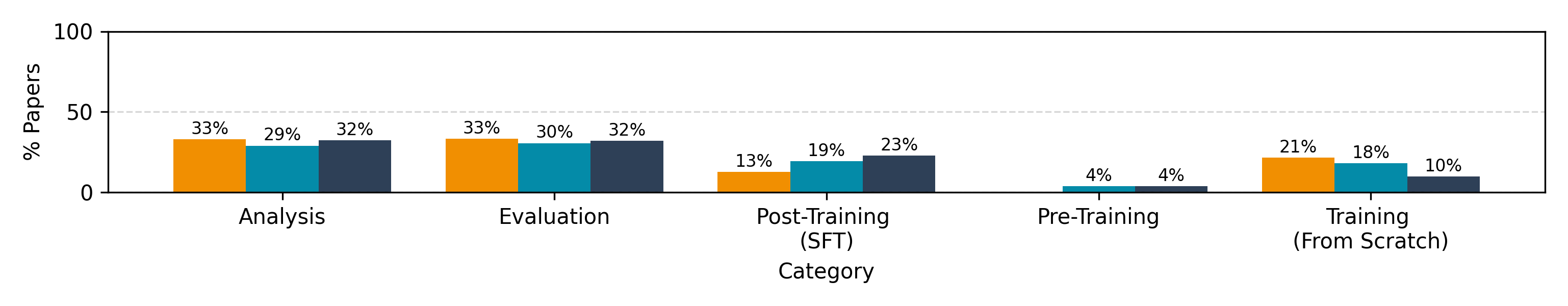}
        \caption{Trends in Task Utility of Papers at Interspeech.}
        \label{fig:task-utility-interspeech}
    \end{subfigure}
    \begin{subfigure}[b]{\textwidth}
        \includegraphics[width=\textwidth]{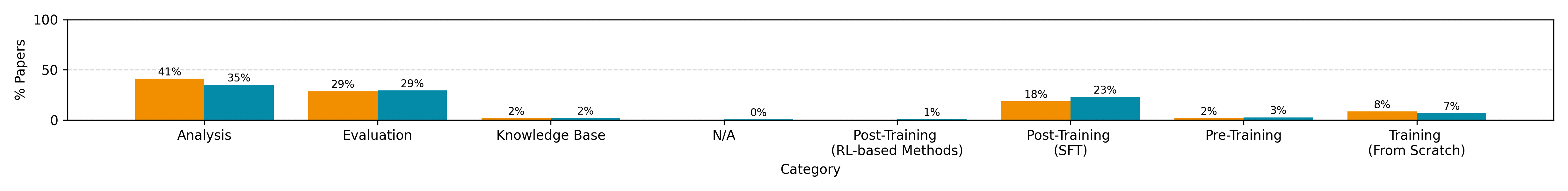}
        \caption{Trends in Task Utility of Papers at LREC.}
        \label{fig:task-utility-lrec}
    \end{subfigure}
    \begin{subfigure}[b]{\textwidth}
        \includegraphics[width=\textwidth]{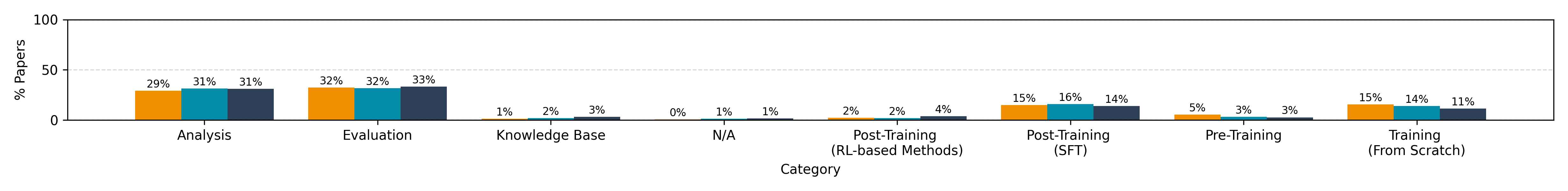}
        \caption{Trends in Task Utility of Papers at NeurIPS.}
        \label{fig:task-utility-neurips}
    \end{subfigure}
    \caption{Paper statistics across conferences on Task Utility at ICML, Interspeech, LREC, and NeurIPS (2022-2024).}
    \label{fig:paper-statistics-task-utility-part2}
\end{figure}

Figures \ref{fig:paper-statistics-task-utility-part1} and \ref{fig:paper-statistics-task-utility-part2} show the paper trends on \textbf{Task Utility} at ACL, CVPR, EMNLP, ICLR, ICML, Interspeech, LREC, and NeurIPS.

\clearpage

\section{Metric Evaluation Rubrics}
\label{appendix:schema}
In this section, we present the template and structured schema used to define each metric dimension in $\methodname$.
\subsection{Data Sources}
\label{subsection:data-sources}
\subsubsection{Template}
\begin{tcolorbox}[colback=gray!10,colframe=black,title=Metric Evaluation Template for Data Sources,breakable]
\#\#\# INSTRUCTION\\
List all of data source entries from only the **new dataset** introduced in the paper, each with one modality and a multi-label origin classification.\\
Please follow these rules:\\
- **Only assess new datasets** that are introduced by the authors. Do **not** evaluate any pre-existing datasets mentioned in the paper.\\
- Base your judgments **strictly on the content of the paper**. Do **not** infer or speculate beyond what is explicitly stated.\\
- Provide clear references and reasoning for each modality and the origin of the data.\\

\#\#\# PAPER\\
\{\{ paper\_text \}\} \\

\#\#\# RESPONSE FORMAT \\
Return a JSON response in the following format:\\

\{\{ format \}\}\\

\#\#\# RESPONSE
\end{tcolorbox}

\subsubsection{Schema}

\begin{tcolorbox}[colback=gray!10,colframe=black,title=Schema for Data Sources,breakable,fontupper=\fontsize{8pt}{8pt}]
\begin{verbatim}
{
  "name": "sources",
  "schema": {
    "type": "object",
    "properties": {
      "sources": {
        "type": "array",
        "description": "List of data source entries, each with one modality and a 
        multi-label origin classification.",
        "items": {
          "type": "object",
          "properties": {
            "Modality": {
              "type": "string",
              "description": "The single data modality present (e.g., 'text', 'image', 
              'audio', 'video', etc.) based on the enum.",
              "enum": ["text", "image", "audio", "video", "graph", "tabular", 
              "time series", "signal/sensor", "other"]
            },
            "Human Generated": {
              "type": "boolean",
              "description": "Set to true if the data (for the specified modality)
              originates from or is captured with human involvement — including
              manual creation, human recording, or data collected through human-operated
              tools such as cameras or sensors."
            },
            "Model Generated": {
              "type": "boolean",
              "description": "Set to true if the data (for the specified modality)
              is generated or simulated by any algorithmic system — including
              AI models, procedural generators, simulations, or other automated
              programmatic methods without direct human authorship."
            },
            "Unknown Origin": {
              "type": "boolean",
              "description": "Set to true if the specified data with the modality 
              mentioned is not specified, not reported, or derived from web-crawl data 
              with unclear provenance."
            },
            "Reference": {
              "type": "string",
              "description": "The location in the paper confirming the previous 
              information provided (e.g., 'Section 1.1')."
            },
            "Reasoning": {
              "type": "string",
              "description": "A justification based on the paper content explaining 
              why the dataset is of modality mentioned, along with the reasoning of 
              its origins."
            }
          },
          "required": [
            "Modality",
            "Human Generated",
            "Model Generated",
            "Unknown Origin",
            "Reference",
            "Reasoning"
          ],
          "additionalProperties": false
        }
      }
    },
    "required": ["sources"],
    "additionalProperties": false
  },
  "strict": true
}
\end{verbatim}
\end{tcolorbox}

\subsection{Data Annotations}
\label{subsection:data-annotations}
\subsubsection{Template}
\begin{tcolorbox}[colback=gray!10,colframe=black,title=Metric Evaluation Template for Data Annotations,breakable]
\#\#\# INSTRUCTION \\
List all of annotation information entries from only the **new dataset** introduced in the paper, each with the annotator and whether instructions, rubrics, and/or examples are present. \\

Follow these strict guidelines: \\
- Only evaluate datasets that are newly introduced in this paper. Do **not** evaluate any pre-existing datasets mentioned in the paper. \\
- Base your judgments **strictly on the content of the paper**. Do **not** infer or speculate beyond what is explicitly stated. \\
- Provide clear references and reasoning for each required annotation information. \\

\#\#\# PAPER \\
\{\{ paper\_text \}\} \\

\#\#\# RESPONSE FORMAT \\
Return a JSON response in the following format: \\

\{\{ format \}\} \\

\#\#\# RESPONSE \\
\end{tcolorbox}

\subsubsection{Schema}

\begin{tcolorbox}[colback=gray!10,colframe=black,title=Schema for Data Annotations,breakable,fontupper=\fontsize{8pt}{8pt}]
\begin{verbatim}
{
  "name": "annotations_data_annotations",
  "schema": {
    "type": "object",
    "properties": {
      "data_annotations": {
        "type": "array",
        "description": "List of annotators along with the data annotation guideline
        evaluations.",
        "items": {
          "type": "object",
          "description": "Each entry assesses who performed the annotation (human or
          model), along with the presence and quality of annotation guidelines
          such as instructions, rubrics, and examples.",
          "properties": {
            "Subject Annotators": {
              "type": "object",
              "description": "Describes who performed the annotations. Choose exactly
              one from a predefined list.",
              "properties": {
                "category": {
                  "type": "string",
                  "description": "Who performed the annotation. Some definitions:
                  \n- Expert: A subject-matter expert or someone from the
                  task's target demographic. 
                  \n- AI Model: An annotation process using LLM or other learned model
                  \n- Automatic Process: A deterministic or simulated method
                  (e.g., rule-based, physics simulation) that is not an AI model.",
                  "enum": [
                    "Single Human Expert",
                    "Multiple Human Experts",
                    "Single Human Non-Expert",
                    "Multiple Human Non-Experts",
                    "AI Model",
                    "Automatic Process"
                  ]
                },
                "reference": {
                  "type": "string",
                  "description": "The location in the paper where this information
                  is discussed (e.g., 'Section 3.1', 'Appendix B')."
                },
                "reasoning": {
                  "type": "string",
                  "description": "Explanation supporting the selected category based
                  on the paper content."
                }
              },
              "required": ["category", "reference", "reasoning"],
              "additionalProperties": false  
            },
            "Has Instructions": {
              "type": "object",
              "description": "Indicates whether detailed annotation instructions are
              provided in the annotation guidelines for the 'Subject'
              mentioned earlier.",
              "properties": {
                "is_applicable": {
                  "type": "boolean",
                  "description": "Set to true if instructions are provided in the
                  annotation guidelines for the 'Subject' mentioned earlier."
                },
                "reference": {
                  "type": "string",
                  "description": "The location in the paper where this is described
                  (e.g., 'Section 3.1')."
                },
                "reasoning": {
                  "type": "string",
                  "description": "Justification from the paper content supporting the
                  inclusion of instructions."
                }
              },
              "required": ["is_applicable", "reference", "reasoning"],
              "additionalProperties": false
            },
            "Has Rubrics": {
              "type": "object",
              "description": "Indicates whether scoring rubrics are provided
              in the annotation guidelines for the 'Subject' mentioned earlier.",
              "properties": {
                "is_applicable": {
                  "type": "boolean",
                  "description": "Set to true if rubrics are provided in the
                  annotation guidelines for the 'Subject' mentioned earlier."
                },
                "reference": {
                  "type": "string",
                  "description": "The location in the paper where this is described
                  (e.g., 'Section 3.1')."
                },
                "reasoning": {
                  "type": "string",
                  "description": "Justification from the paper content supporting
                  rubric usage."
                }
              },
              "required": ["is_applicable", "reference", "reasoning"],
              "additionalProperties": false
            },
            "Has Examples": {
              "type": "object",
              "description": "Indicates whether annotation examples are provided
              in the annotation guidelines for the 'Subject' mentioned earlier.",
              "properties": {
                "is_applicable": {
                  "type": "boolean",
                  "description": "Set to true if examples are provided in the
                  annotation guidelines for the 'Subject' mentioned earlier."
                },
                "reference": {
                  "type": "string",
                  "description": "The location in the paper where this is described
                  (e.g., 'Appendix B')."
                },
                "reasoning": {
                  "type": "string",
                  "description": "Justification from the paper content supporting
                  the inclusion of examples."
                }
              },
              "required": ["is_applicable", "reference", "reasoning"],
              "additionalProperties": false
            }
          },
          "required": [
            "Subject Annotators",
            "Has Instructions",
            "Has Rubrics",
            "Has Examples"
          ],
          "additionalProperties": false
        }
      }
    },
    "required": ["data_annotations"],
    "additionalProperties": false
  },
  "strict": true
}
\end{verbatim}
\end{tcolorbox}

\subsection{Quality Assurance}
\label{subsection:quality-assurance}

\subsubsection{Template}
\begin{tcolorbox}[colback=gray!10,colframe=black,title=Metric Evaluation Template for Quality Assurance,breakable]
\#\#\# INSTRUCTION\\
Carefully evaluate the quality and characteristics of the **new datasets** introduced in the paper using the rubric provided below. \\

Please follow these rules: \\
- **Only assess new datasets** that are introduced by the authors. Do **not** evaluate any pre-existing datasets mentioned in the paper. \\
- Base your judgments **strictly on the content of the paper**. Do **not** infer or speculate beyond what is explicitly stated. \\
- Use the rubric definitions to guide your labeling. Provide clear references and reasoning wherever applicable. \\

\#\#\# PAPER\\
\{\{ paper\_text \}\} \\

\#\#\# RUBRIC \\
Metric name: Quality Assurance \\
Metric description: This metric assesses the rigor and reliability of the quality assurance (QA) process used to validate dataset annotations or content. It captures whether QA was performed by experts, non-experts, machines, or not at all, and whether the process is transparently reported. This helps determine the overall trustworthiness and consistency of the data. \\
Multiple labels are allowed, except for mutually exclusive categories. Specifically: \\
    - 'Single Human Expert' and 'Multiple Human Experts' cannot be both true. \\
    - 'Single Human Non-Expert' and 'Multiple Human Non-Experts' cannot be both true. \\
    - 'N/A' is a single-label option and cannot be combined with others. \\
Each label indicates whether it applies to the dataset, with supporting evidence from the paper. \\

Options: \\
Single Human Expert: Quality assurance is conducted by a single human annotator who is either a subject matter expert or a member of the target demographic. If there is no information about the annotator, then the annotator is not an expert. \\
Multiple Human Experts: Quality assurance is performed by multiple human annotators with subject matter expertise or belonging to the target demographic. If there is no information about the annotators, then the annotators are not experts. \\
Single Human Non-Expert": Quality assurance is conducted by a single human annotator without subject matter expertise.\\
Multiple Human Non-Experts": Quality assurance is conducted by multiple human annotators who do not possess subject matter expertise. \\
Automatic Process: Quality assurance is conducted through the automated verification of code or formulas using algorithmic or rule-based techniques. \\
AI Model: Quality assurance is performed by an AI model as a judge. \\
N/A: No quality assurance process is applied, or none is documented. \\

\#\#\# RESPONSE FORMAT \\
Return a JSON response in the following format:\\

\{\{ format \}\}\\

\#\#\# RESPONSE
\end{tcolorbox}

\subsubsection{Schema}

\begin{tcolorbox}[colback=gray!10,colframe=black,title=Schema for Quality Assurance,breakable,fontupper=\fontsize{8pt}{8pt}]
\begin{lstlisting}[basicstyle=\fontsize{8pt}{8pt}\ttfamily,
    breaklines=true,
    breakatwhitespace=true,
    showstringspaces=false,
    % language=json,
    frame=none,
    numbers=none
    ]
{
  "name": "annotations_quality_assurance",
  "schema": {
    "type": "object",
    "properties": {
      "quality_assurance": {
        "type": "object",
        "description": "This metric assesses the rigor and reliability of the quality assurance (QA) process used to validate dataset annotations or content. It captures whether QA was performed by experts, non-experts, machines, or not at all, and whether the process is transparently reported.",
        "properties": {
          "Single Human Expert": {
            "type": "object",
            "description": "Quality assurance is conducted by a single human annotator who is either a subject matter expert or a member of the target demographic. If there is no information about the annotator, then the annotator is not an expert.",
            "properties": {
              "is_applicable": {
                "type": "boolean",
                "description": "Set to true if this label applies to the dataset. If true, reference and reasoning must be provided."
              },
              "reference": {
                "type": "string",
                "description": "The location in the paper where this is described (e.g., 'Section 3.1')."
              },
              "reasoning": {
                "type": "string",
                "description": "Justification based on the paper content explaining why the QA was performed by a single human expert."
              }
            },
            "required": ["is_applicable", "reference", "reasoning"],
            "additionalProperties": false
          },
          "Multiple Human Experts": {
            "type": "object",
            "description": "Quality assurance is performed by multiple human annotators with subject matter expertise or belonging to the target demographic. If there is no information about the annotators, then the annotators are not experts.",
            "properties": {
              "is_applicable": {
                "type": "boolean",
                "description": "Set to true if this label applies to the dataset. If true, reference and reasoning must be provided."
              },
              "reference": {
                "type": "string",
                "description": "The location in the paper where this is described (e.g., 'Section 3.2')."
              },
              "reasoning": {
                "type": "string",
                "description": "Justification based on the paper content explaining why the QA involved multiple human experts."
              }
            },
            "required": ["is_applicable", "reference", "reasoning"],
            "additionalProperties": false
          },
          "Single Human Non-Expert": {
            "type": "object",
            "description": "Quality assurance is conducted by a single human annotator without subject matter expertise.",
            "properties": {
              "is_applicable": {
                "type": "boolean",
                "description": "Set to true if this label applies to the dataset. If true, reference and reasoning must be provided."
              },
              "reference": {
                "type": "string",
                "description": "The location in the paper where this is described (e.g., 'Section 3.3')."
              },
              "reasoning": {
                "type": "string",
                "description": "Justification based on the paper content explaining why the QA was done by a single non-expert."
              }
            },
            "required": ["is_applicable", "reference", "reasoning"],
            "additionalProperties": false
          },
          "Multiple Human Non-Experts": {
            "type": "object",
            "description": "Quality assurance is conducted by multiple human annotators who do not possess subject matter expertise.",
            "properties": {
              "is_applicable": {
                "type": "boolean",
                "description": "Set to true if this label applies to the dataset. If true, reference and reasoning must be provided."
              },
              "reference": {
                "type": "string",
                "description": "The location in the paper where this is described (e.g., 'Section 3.4')."
              },
              "reasoning": {
                "type": "string",
                "description": "Justification based on the paper content explaining why the QA involved multiple non-expert annotators."
              }
            },
            "required": ["is_applicable", "reference", "reasoning"],
            "additionalProperties": false
          },
          "AI Model": {
            "type": "object",
            "description": "Quality assurance is performed by an AI model as a judge.",
            "properties": {
              "is_applicable": {
                "type": "boolean",
                "description": "Set to true if this label applies to the dataset. If true, reference and reasoning must be provided."
              },
              "reference": {
                "type": "string",
                "description": "The location in the paper where this is described (e.g., 'Section 3.6')."
              },
              "reasoning": {
                "type": "string",
                "description": "Justification based on the paper content explaining why an AI model was used for QA."
              }
            },
            "required": ["is_applicable", "reference", "reasoning"],
            "additionalProperties": false
          },
          "Automatic Process": {
            "type": "object",
            "description": "Quality assurance is conducted through the automated verification of code or formulas using algorithmic or rule-based techniques.",
            "properties": {
              "is_applicable": {
                "type": "boolean",
                "description": "Set to true if this label applies to the dataset. If true, reference and reasoning must be provided."
              },
              "reference": {
                "type": "string",
                "description": "The location in the paper where this is described (e.g., 'Section 3.5')."
              },
              "reasoning": {
                "type": "string",
                "description": "Justification based on the paper content explaining why the QA is considered automated verification."
              }
            },
            "required": ["is_applicable", "reference", "reasoning"],
            "additionalProperties": false
          },
          "N/A": {
            "type": "object",
            "description": "No quality assurance process is applied, or none is documented.",
            "properties": {
              "is_applicable": {
                "type": "boolean",
                "description": "Set to true if no quality assurance is described or performed. If true, reasoning must be provided."
              },
              "reasoning": {
                "type": "string",
                "description": "Justification based on the paper content explaining why no QA process is considered present."
              }
            },
            "required": ["is_applicable", "reasoning"],
            "additionalProperties": false
          }
        },
        "required": [
          "Single Human Expert",
          "Multiple Human Experts",
          "Single Human Non-Expert",
          "Multiple Human Non-Experts",
          "AI Model",
          "Automatic Verification",
          "N/A"
        ],
        "additionalProperties": false
      }
    },
    "required": ["quality_assurance"],
    "additionalProperties": false
  },
  "strict": true
}

\end{lstlisting}
\end{tcolorbox}

\subsection{Data Novelty}
\label{subsection:data-novelty}
\subsubsection{Template}
\begin{tcolorbox}[colback=gray!10,colframe=black,title=Metric Evaluation Template for Data Novelty,breakable]
\#\#\# INSTRUCTION\\
Carefully evaluate the quality and characteristics of the **new datasets** introduced in the paper using the rubric provided below. \\

Please follow these rules: \\
- **Only assess new datasets** that are introduced by the authors. Do **not** evaluate any pre-existing datasets mentioned in the paper. \\
- Base your judgments **strictly on the content of the paper**. Do **not** infer or speculate beyond what is explicitly stated. \\
- Use the rubric definitions to guide your labeling. Provide clear references and reasoning wherever applicable. \\
    
\#\#\# PAPER \\
\{\{ paper\_text \}\} \\

\#\#\# RUBRIC \\
Metric: Data Novelty \\ 
Description: This metric assesses the originality of the data based on how it was generated. Multiple labels may be selected, except for 'N/A', which is exclusive and cannot be combined with other options. Each label indicates whether it applies to the dataset, with supporting evidence from the paper. \\

Options: \\
New Data from Human: Original content created entirely from scratch by human contributors. It is not translated, adapted, or derived from pre-existing material.\\
New Data from Model: Original content generated entirely by AI or machine learning models without reference to or transformation of existing data.\\
Human Translation: Data produced by translating content from another language through human translators.\\
Machine Translation: Data generated by translating content from another language using machine translation systems.\\
Collated: Data collected or aggregated from existing sources without significant modification.\\
Derived: Data based on existing sources, with some modifications, transformations, or adaptations applied.\\
N/A: The data source or method of generation is not specified or documented.\\

\#\#\# RESPONSE FORMAT \\
Return a JSON response in the following format:\\

\{\{ format \}\}\\

\#\#\# RESPONSE
\end{tcolorbox}

\subsubsection{Schema}

\begin{tcolorbox}[colback=gray!10,colframe=black,title=Schema for Data Novelty,breakable,fontupper=\fontsize{8pt}{8pt}]
\begin{lstlisting}[basicstyle=\fontsize{8pt}{8pt}\ttfamily,
    breaklines=true,
    breakatwhitespace=true,
    showstringspaces=false,
    % language=json,
    frame=none,
    numbers=none
    ]
{
    "name": "utility_data_novelty",
    "schema": {
      "type": "object",
      "properties": {
        "data_novelty": {
          "type": "object",
          "description": "This metric assesses the originality of the data based on how it was generated.",
          "properties": {
            "New Data from Human": {
              "type": "object",
              "description": "Original content created entirely from scratch by human contributors. It is not translated, adapted, or derived from pre-existing material.",
              "properties": {
                "is_applicable": {
                  "type": "boolean",
                  "description": "Set to true if this applies to the dataset. If true, reference and reasoning must be provided."
                },
                "reference": {
                  "type": "string",
                  "description": "Location in the paper where this is stated (e.g., 'Section 2.1')."
                },
                "reasoning": {
                  "type": "string",
                  "description": "Explanation based on the paper content justifying why the data is considered newly created by humans."
                }
              },
              "required": ["is_applicable", "reference", "reasoning"],
              "additionalProperties": false
            },
            "New Data from Model": {
              "type": "object",
              "description": "Original content generated entirely by AI or machine learning models without reference to or transformation of existing data.",
              "properties": {
                "is_applicable": {
                  "type": "boolean",
                  "description": "Set to true if this applies to the dataset. If true, reference and reasoning must be provided."
                },
                "reference": {
                  "type": "string",
                  "description": "Location in the paper where this is stated (e.g., 'Section 2.2')."
                },
                "reasoning": {
                  "type": "string",
                  "description": "Explanation based on the paper content justifying why the data is considered newly generated by a model."
                }
              },
              "required": ["is_applicable", "reference", "reasoning"],
              "additionalProperties": false
            },
            "Human Translation": {
              "type": "object",
              "description": "Data produced by translating content from another language through human translators.",
              "properties": {
                "is_applicable": {
                  "type": "boolean",
                  "description": "Set to true if this applies to the dataset. If true, reference and reasoning must be provided."
                },
                "reference": {
                  "type": "string",
                  "description": "Location in the paper where this is stated (e.g., 'Section 3.1')."
                },
                "reasoning": {
                  "type": "string",
                  "description": "Explanation based on the paper content justifying why the data is considered human-translated."
                }
              },
              "required": ["is_applicable", "reference", "reasoning"],
              "additionalProperties": false
            },
            "Machine Translation": {
              "type": "object",
              "description": "Data generated by translating content from another language using machine translation systems.",
              "properties": {
                "is_applicable": {
                  "type": "boolean",
                  "description": "Set to true if this applies to the dataset. If true, reference and reasoning must be provided."
                },
                "reference": {
                  "type": "string",
                  "description": "Location in the paper where this is stated (e.g., 'Section 3.2')."
                },
                "reasoning": {
                  "type": "string",
                  "description": "Explanation based on the paper content justifying why the data is considered machine-translated."
                }
              },
              "required": ["is_applicable", "reference", "reasoning"],
              "additionalProperties": false
            },
            "Collated": {
              "type": "object",
              "description": "Data collected or aggregated from existing sources without significant modification.",
              "properties": {
                "is_applicable": {
                  "type": "boolean",
                  "description": "Set to true if this applies to the dataset. If true, reference and reasoning must be provided."
                },
                "reference": {
                  "type": "string",
                  "description": "Location in the paper where this is stated (e.g., 'Section 4.1')."
                },
                "reasoning": {
                  "type": "string",
                  "description": "Explanation based on the paper content justifying why the data is considered collated."
                }
              },
              "required": ["is_applicable", "reference", "reasoning"],
              "additionalProperties": false
            },
            "Derived": {
              "type": "object",
              "description": "Data based on existing sources, with some modifications, transformations, or adaptations applied.",
              "properties": {
                "is_applicable": {
                  "type": "boolean",
                  "description": "Set to true if this applies to the dataset. If true, reference and reasoning must be provided."
                },
                "reference": {
                  "type": "string",
                  "description": "Location in the paper where this is stated (e.g., 'Section 4.2')."
                },
                "reasoning": {
                  "type": "string",
                  "description": "Explanation based on the paper content justifying why the data is considered derived."
                }
              },
              "required": ["is_applicable", "reference", "reasoning"],
              "additionalProperties": false
            },
            "N/A": {
              "type": "object",
              "description": "The data source or method of generation is not specified or documented.",
              "properties": {
                "is_applicable": {
                  "type": "boolean",
                  "description": "Set to true only if no other categories apply and the origin is not documented."
                },
                "reasoning": {
                  "type": "string",
                  "description": "Explanation based on the paper content justifying why the data origin is considered unknown."
                }
              },
              "required": ["is_applicable", "reasoning"],
              "additionalProperties": false
            }
          },
          "required": [
            "New Data from Human",
            "New Data from Model",
            "Human Translation",
            "Machine Translation",
            "Collated",
            "Derived",
            "N/A"
          ],
          "additionalProperties": false
        }
      },
      "required": ["data_novelty"],
      "additionalProperties": false
    },
    "strict": true
  }
  
\end{lstlisting}
\end{tcolorbox}

\subsection{Task Utility}
\label{subsection:data-utility}

\subsubsection{Template}
\begin{tcolorbox}[colback=gray!10,colframe=black,title=Metric Evaluation Template for Task Utility,breakable]
\#\#\# INSTRUCTION\\
Carefully evaluate the quality and characteristics of the **new datasets** introduced in the paper using the rubric provided below. \\

Please follow these rules: \\
- **Only assess new datasets** that are introduced by the authors. Do **not** evaluate any pre-existing datasets mentioned in the paper. \\
- Base your judgments **strictly on the content of the paper**. Do **not** infer or speculate beyond what is explicitly stated. \\
- Use the rubric definitions to guide your labeling. Provide clear references and reasoning wherever applicable. \\

\#\#\# PAPER\\
\{\{ paper\_text \}\} \\

\#\#\# RUBRIC \\
Metric: Task Utility \\
Description: This metric identifies how the dataset is used within the machine learning pipeline. Understanding its utility helps clarify the dataset's purpose, relevance, and integration into model development or evaluation workflows. Multiple labels may be selected, except for 'N/A', which is exclusive and cannot be combined with other options. Each label indicates whether it applies to the dataset, with supporting evidence from the paper. \\

Options: \\
Pre-Training: The proposed dataset in the paper is used exclusively for pre-training large models on general patterns, typically in an unsupervised or self-supervised manner. \\
Post-Training (Supervised Fine-tuning): The proposed dataset in the paper is used to fine-tune a pre-trained model using supervised learning methods. \\
Post-Training (RL-based Methods): The proposed dataset in the paper is used for reinforcement learning post-training techniques such as RLHF. \\
Evaluation: The proposed dataset in the paper is used exclusively for evaluation, benchmarking, or performance measurement. \\
Analysis: The proposed dataset in the paper is used primarily for analyzing trends, patterns, or characteristics rather than training or evaluation. \\
Knowledge Base: The proposed dataset in the paper serves as a knowledge base to augment models (e.g., through retrieval-augmented generation). \\
N/A: No practical usage of The proposed dataset in the paper is described or demonstrated in the paper. \\

\#\#\# RESPONSE FORMAT \\
Return a JSON response in the following format:\\

\{\{ format \}\}\\

\#\#\# RESPONSE
\end{tcolorbox}

\subsubsection{Schema}

\begin{tcolorbox}[colback=gray!10,colframe=black,title=Schema for Task Utility,breakable,fontupper=\fontsize{8pt}{8pt}]
\begin{lstlisting}[basicstyle=\fontsize{8pt}{8pt}\ttfamily,
    breaklines=true,
    breakatwhitespace=true,
    showstringspaces=false,
    % language=json,
    frame=none,
    numbers=none
    ]
{
  "name": "utility_task_utility",
  "schema": {
    "type": "object",
    "properties": {
      "task_utility": {
        "type": "object",
        "description": "This metric identifies how the dataset is used within the machine learning pipeline. Understanding its utility helps clarify the dataset's purpose, relevance, and integration into model development or evaluation workflows.",
        "properties": {
          "Pre-Training": {
            "type": "object",
            "description": "The dataset is used exclusively for pre-training large models on general patterns, typically in an unsupervised or self-supervised manner.",
            "properties": {
              "is_applicable": {
                "type": "boolean",
                "description": "Set to true if the dataset is used for pre-training. If true, reference and reasoning must be provided."
              },
              "reference": {
                "type": "string",
                "description": "Location in the paper where this usage is described (e.g., 'Section 2.1')."
              },
              "reasoning": {
                "type": "string",
                "description": "Justification from the paper explaining why the dataset is used for pre-training."
              }
            },
            "required": ["is_applicable", "reference", "reasoning"],
            "additionalProperties": false
          },
          "Training (From Scratch)": {
            "type": "object",
            "description": "The dataset is used to train a model from randomly initialized parameters (i.e., not pre-trained).",
            "properties": {
              "is_applicable": {
                "type": "boolean",
                "description": "Set to true if the dataset is used for training from scratch. If true, reference and reasoning must be provided."
              },
              "reference": {
                "type": "string",
                "description": "Location in the paper where this usage is described (e.g., 'Section 2.1')."
              },
              "reasoning": {
                "type": "string",
                "description": "Justification from the paper explaining why the dataset is used for training from scratch."
              }
            },
            "required": ["is_applicable", "reference", "reasoning"],
            "additionalProperties": false
          },
          "Post-Training (Supervised Fine-tuning)": {
            "type": "object",
            "description": "The dataset is used to fine-tune a pre-trained model using supervised learning methods.",
            "properties": {
              "is_applicable": {
                "type": "boolean",
                "description": "Set to true if the dataset is used for supervised fine-tuning. If true, reference and reasoning must be provided."
              },
              "reference": {
                "type": "string",
                "description": "Location in the paper where this usage is described (e.g., 'Section 3.1')."
              },
              "reasoning": {
                "type": "string",
                "description": "Justification from the paper explaining why the dataset is used for supervised fine-tuning."
              }
            },
            "required": ["is_applicable", "reference", "reasoning"],
            "additionalProperties": false
          },
          "Post-Training (RL-based Methods)": {
            "type": "object",
            "description": "The dataset is used for reinforcement learning post-training techniques such as RLHF.",
            "properties": {
              "is_applicable": {
                "type": "boolean",
                "description": "Set to true if the dataset is used in RL-based post-training. If true, reference and reasoning must be provided."
              },
              "reference": {
                "type": "string",
                "description": "Location in the paper where this usage is described (e.g., 'Section 3.3')."
              },
              "reasoning": {
                "type": "string",
                "description": "Justification from the paper explaining why the dataset is used in RL-based methods."
              }
            },
            "required": ["is_applicable", "reference", "reasoning"],
            "additionalProperties": false
          },
          "Evaluation": {
            "type": "object",
            "description": "The dataset is used exclusively for evaluation, benchmarking, or performance measurement.",
            "properties": {
              "is_applicable": {
                "type": "boolean",
                "description": "Set to true if the dataset is used for evaluation. If true, reference and reasoning must be provided."
              },
              "reference": {
                "type": "string",
                "description": "Location in the paper where this usage is described (e.g., 'Section 4.1')."
              },
              "reasoning": {
                "type": "string",
                "description": "Justification from the paper explaining why the dataset is used for evaluation."
              }
            },
            "required": ["is_applicable", "reference", "reasoning"],
            "additionalProperties": false
          },
          "Analysis": {
            "type": "object",
            "description": "The dataset is used primarily for analyzing trends, patterns, or characteristics rather than training or evaluation.",
            "properties": {
              "is_applicable": {
                "type": "boolean",
                "description": "Set to true if the dataset is used for analysis. If true, reference and reasoning must be provided."
              },
              "reference": {
                "type": "string",
                "description": "Location in the paper where this usage is described (e.g., 'Section 5.1')."
              },
              "reasoning": {
                "type": "string",
                "description": "Justification from the paper explaining why the dataset is used for analysis."
              }
            },
            "required": ["is_applicable", "reference", "reasoning"],
            "additionalProperties": false
          },
          "Knowledge Base": {
            "type": "object",
            "description": "The dataset serves as a knowledge base to augment models (e.g., through retrieval-augmented generation).",
            "properties": {
              "is_applicable": {
                "type": "boolean",
                "description": "Set to true if the dataset is used as a knowledge base.
                If true, reference and reasoning must be provided."
              },
              "reference": {
                "type": "string",
                "description": "Location in the paper where this usage is described (e.g., 'Section 6.1')."
              },
              "reasoning": {
                "type": "string",
                "description": "Justification from the paper explaining why 
                the dataset is considered a knowledge base."
              }
            },
            "required": ["is_applicable", "reference", "reasoning"],
            "additionalProperties": false
          },
          "N/A": {
            "type": "object",
            "description": "No practical usage of the dataset is described or demonstrated in the paper.",
            "properties": {
              "is_applicable": {
                "type": "boolean",
                "description": "Set to true only if no other categories
                apply and there is no documented usage."
              },
              "reasoning": {
                "type": "string",
                "description": "Explanation justifying why no
                documented utility is found in the paper."
              }
            },
            "required": ["is_applicable", "reasoning"],
            "additionalProperties": false
          }
        },
        "required": [
          "Pre-Training",
          "Training (From Scratch)",
          "Post-Training (Supervised Fine-tuning)",
          "Post-Training (RL-based Methods)",
          "Evaluation",
          "Analysis",
          "Knowledge Base",
          "N/A"
        ],
        "additionalProperties": false
      }
    },
    "required": ["task_utility"],
    "additionalProperties": false
  },
  "strict": true
}

\end{lstlisting}
\end{tcolorbox}

\subsection{Language Coverage}
\label{subsection:lang-coverage}

\subsubsection{Template}
\begin{tcolorbox}[colback=gray!10,colframe=black,title=Metric Evaluation Template for Language Coverage,breakable]
\#\#\# INSTRUCTION\\
Carefully evaluate the quality and characteristics of the **new datasets** introduced in the paper using the rubric provided below. \\

Please follow these rules: \\
- **Only assess new datasets** that are introduced by the authors. Do **not** evaluate any pre-existing datasets mentioned in the paper. \\
- Base your judgments **strictly on the content of the paper**. Do **not** infer or speculate beyond what is explicitly stated. \\
- Use the rubric definitions to guide your labeling. Provide clear references and reasoning wherever applicable. \\

\#\#\# PAPER\\
\{\{ paper\_text \}\} \\

\#\#\# RUBRIC\\
Metric: Language Coverage \\
Description: This metric categorizes the linguistic scope of the dataset, indicating how many and which types of languages are included. This helps assess the dataset's applicability to multilingual or language-specific research. This metric allows multiple labels, except for 'Unknown Origin' and 'N/A', which must be used as single, exclusive labels. Each label indicates whether it applies to the dataset, with supporting evidence from the paper. \\

Options: \\
Multilingual: The proposed dataset in the paper contains entries with more than two human languages. \\
Bilingual: The proposed dataset in the paper contains entries with exactly two human languages. \\
Monolingual (English): The proposed dataset in the paper contains entries with only English content. \\
Monolingual (Non-English): The proposed dataset in the paper contains entries with exactly one language that is non-English. \\
Code / Programming Language: The proposed dataset in the paper contains entries with programming or structured code-related content (e.g., Python, HTML, SQL, bytecode). \\
Mathematical and Logical Notation: "The proposed dataset in the paper contains entries with mathematical or formal logical expressions or symbolic representations. \\
Biological and Non-Human Communication Systems: "The proposed dataset in the paper contains entries with biological sequences or non-human communication (e.g., DNA, animal signals, chemical signaling). \\
Constructed Language: The proposed dataset in the paper contains entries with fictional or artificially created languages such as Klingon or Esperanto. \\
Unknown: The proposed dataset in the paper contains entries with language(s) listed previously but the language(s) are not specified or documented. \\
N/A: The proposed dataset in the paper does not contain entries with any languages. \\

\#\#\# RESPONSE FORMAT \\
Return a JSON response in the following format:\\

\{\{ format \}\}\\

\#\#\# RESPONSE
\end{tcolorbox}

\subsubsection{Schema}
\begin{tcolorbox}[colback=gray!10,colframe=black,title=Schema for Task Utility,breakable,fontupper=\fontsize{8pt}{8pt}]
\begin{lstlisting}[basicstyle=\fontsize{8pt}{8pt}\ttfamily,
    breaklines=true,
    breakatwhitespace=true,
    showstringspaces=false,
    % language=json,
    frame=none,
    numbers=none
    ]
{
  "name": "utility_lang",
  "schema": {
    "type": "object",
    "properties": {
      "lang": {
        "type": "object",
        "description": "This metric categorizes the linguistic scope of the proposed dataset in the paper, indicating how many and which types of languages are included.",
        "properties": {
          "Multilingual": {
            "type": "object",
            "description": "The proposed dataset in the paper contains entries with more than two human languages.",
            "properties": {
              "is_applicable": {
                "type": "boolean",
                "description": "Set to true if the description applies to the dataset. If true, reference and reasoning must be provided."
              },
              "reference": {
                "type": "string",
                "description": "Location in the paper where this is stated (e.g., 'Section 2.1')."
              },
              "reasoning": {
                "type": "string",
                "description": "Explanation based on the paper content justifying why the data is considered multilingual along with the list of languages used."
              }
            },
            "required": ["is_applicable", "reference", "reasoning"],
            "additionalProperties": false
          },
          "Bilingual": {
            "type": "object",
            "description": "The proposed dataset in the paper contains entries with exactly two human languages.",
            "properties": {
              "is_applicable": {
                "type": "boolean",
                "description": "Set to true if the description applies to the dataset. If true, reference and reasoning must be provided."
              },
              "reference": {
                "type": "string",
                "description": "Location in the paper where this is stated (e.g., 'Section 2.2')."
              },
              "reasoning": {
                "type": "string",
                "description": "Explanation based on the paper content justifying why the data is considered bilingual and mention the two langauges."
              }
            },
            "required": ["is_applicable", "reference", "reasoning"],
            "additionalProperties": false
          },
          "Monolingual (English)": {
            "type": "object",
            "description": "The proposed dataset in the paper contains entries with only English content.",
            "properties": {
              "is_applicable": {
                "type": "boolean",
                "description": "Set to true if the description applies to the dataset. If true, reference and reasoning must be provided."
              },
              "reference": {
                "type": "string",
                "description": "Location in the paper where this is stated (e.g., 'Section 3.1')."
              },
              "reasoning": {
                "type": "string",
                "description": "Explanation based on the paper content justifying why the data is considered monolingual (English)."
              }
            },
            "required": ["is_applicable", "reference", "reasoning"],
            "additionalProperties": false
          },
          "Monolingual (Non-English)": {
            "type": "object",
            "description": "The proposed dataset in the paper contains entries with exactly one language that is non-English.",
            "properties": {
              "is_applicable": {
                "type": "boolean",
                "description": "Set to true if the description applies to the dataset. If true, reference and reasoning must be provided."
              },
              "reference": {
                "type": "string",
                "description": "Location in the paper where this is stated (e.g., 'Section 3.2')."
              },
              "reasoning": {
                "type": "string",
                "description": "Explanation based on the paper content justifying why the data is considered monolingual (non-English) and mention the langauge."
              }
            },
            "required": ["is_applicable", "reference", "reasoning"],
            "additionalProperties": false
          },
          "Code / Programming Language": {
            "type": "object",
            "description": "The proposed dataset in the paper contains entries with programming or structured code-related content (e.g., Python, HTML, SQL, bytecode).",
            "properties": {
              "is_applicable": {
                "type": "boolean",
                "description": "Set to true if the description applies to the dataset. If true, reference and reasoning must be provided."
              },
              "reference": {
                "type": "string",
                "description": "Location in the paper describing this content (e.g., 'Section 3.2')."
              },
              "reasoning": {
                "type": "string",
                "description": "Justification from the paper indicating that code or programming language is present along with their descriptions."
              }
            },
            "required": ["is_applicable", "reference", "reasoning"],
            "additionalProperties": false
          },
          "Mathematical and Logical Notation": {
            "type": "object",
            "description": "The proposed dataset in the paper contains entries with mathematical or formal logical expressions or symbolic representations.",
            "properties": {
              "is_applicable": {
                "type": "boolean",
                "description": "Set to true if the description appliest. If true, reference and reasoning must be provided."
              },
              "reference": {
                "type": "string",
                "description": "Location in the paper describing this content (e.g., 'Section 2.4')."
              },
              "reasoning": {
                "type": "string",
                "description": "Justification from the paper indicating the presence of math or logic notation along with their descriptions."
              }
            },
            "required": ["is_applicable", "reference", "reasoning"],
            "additionalProperties": false
          },
          "Biological and Non-Human Communication Systems": {
            "type": "object",
            "description": "The proposed dataset in the paper contains entries with biological sequences or non-human communication (e.g., DNA, animal signals, chemical signaling).",
            "properties": {
              "is_applicable": {
                "type": "boolean",
                "description": "Set to true if the description applies to the dataset. If true, reference and reasoning must be provided."
              },
              "reference": {
                "type": "string",
                "description": "Location in the paper describing this content (e.g., 'Section 5.1')."
              },
              "reasoning": {
                "type": "string",
                "description": "Justification from the paper explaining why biological or non-human communication data is included along with their descriptions."
              }
            },
            "required": ["is_applicable", "reference", "reasoning"],
            "additionalProperties": false
          },
          "Constructed Language": {
            "type": "object",
            "description": "The proposed dataset in the paper includes fictional or artificially created languages such as Klingon or Esperanto.",
            "properties": {
              "is_applicable": {
                "type": "boolean",
                "description": "Set to true if the description applies to the dataset. If true, reference and reasoning must be provided."
              },
              "reference": {
                "type": "string",
                "description": "Location in the paper describing this content (e.g., 'Section 6.2')."
              },
              "reasoning": {
                "type": "string",
                "description": "Justification from the paper indicating constructed languages are present along with their descriptions."
              }
            },
            "required": ["is_applicable", "reference", "reasoning"],
            "additionalProperties": false
          },
          "Unknown": {
            "type": "object",
            "description": "The entries of the proposed dataset in the paper contain some language(s) but they are not specified or documented.",
            "properties": {
              "is_applicable": {
                "type": "boolean",
                "description": "Set to true if the description applies to the dataset. If true, reference and reasoning must be provided."
              },
              "reference": {
                "type": "string",
                "description": "Location in the paper where this is stated (e.g., 'Section 3.2')."
              },
              "reasoning": {
                "type": "string",
                "description": "Explanation based on the paper content justifying why the language of dataset is considered unknown."
              }
            },
            "required": ["is_applicable", "reference", "reasoning"],
            "additionalProperties": false
          },
          "N/A": {
            "type": "object",
            "description": "The entries of the proposed dataset in the paper does not contain any language.",
            "properties": {
              "is_applicable": {
                "type": "boolean",
                "description": "Set to true only if the dataset does not contain any language."
              },
              "reasoning": {
                "type": "string",
                "description": "Explanation based on the paper content justifying why the data does not contain any language."
              }
            },
            "required": ["is_applicable", "reasoning"],
            "additionalProperties": false
          }
        },
        "required": [
          "Multilingual",
          "Bilingual",
          "Monolingual (English)",
          "Monolingual (Non-English)",
          "Code / Programming Language",
          "Mathematical and Logical Notation",
          "Biological and Non-Human Communication Systems",
          "Constructed Language",
          "Unknown",
          "N/A"
        ],
        "additionalProperties": false
      }
    },
    "required": ["lang"],
    "additionalProperties": false
  },
  "strict": true
}


\end{lstlisting}
\end{tcolorbox}

\subsection{Reproducibility}
\label{subsection:reproducibility}

\subsubsection{Template}
\begin{tcolorbox}[colback=gray!10,colframe=black,title=Metric Evaluation Template for Reproducibility,breakable]
\#\#\# INSTRUCTION\\
Carefully evaluate the quality and characteristics of the **new datasets** introduced in the paper using the rubric provided below. \\

Please follow these rules: \\
- **Only assess new datasets** that are introduced by the authors. Do **not** evaluate any pre-existing datasets mentioned in the paper. \\
- Base your judgments **strictly on the content of the paper**. Do **not** infer or speculate beyond what is explicitly stated. \\
- Use the rubric definitions to guide your labeling. Provide clear references and reasoning wherever applicable. \\

\#\#\# PAPER\\
\{\{ paper\_text \}\} \\

\#\#\# RUBRIC \\
Metric: Reproducibility \\
Description: This metric pertains to whether the code used for constructing the dataset is made publicly available or not for reproducibility and evaluates the transparency and completeness of the dataset creation documentation, which is crucial for reproducibility, ethical assessment, and downstream usability. \\

\#\#\# RESPONSE FORMAT \\
Return a JSON response in the following format:\\

\{\{ format \}\}\\

\#\#\# RESPONSE
\end{tcolorbox}

\subsubsection{Schema}

\begin{tcolorbox}[colback=gray!10,colframe=black,title=Schema for Code Provided,breakable,fontupper=\fontsize{8pt}{8pt}]
\begin{lstlisting}[basicstyle=\fontsize{8pt}{8pt}\ttfamily,
    breaklines=true,
    breakatwhitespace=true,
    showstringspaces=false,
    % language=json,
    frame=none,
    numbers=none
    ]
{
    "name": "utility_reproducibility",
    "schema": {
        "type": "object",
        "properties": {
            "reproducibility": {
                "type": "object",
                "description": "This metric pertains to whether the code used for constructing the dataset is made publicly available or not for reproducibility and evaluates the transparency and completeness of the dataset creation documentation, which is crucial for reproducibility, ethical assessment, and downstream usability.",
                "properties": {
                    "code": {
                        "type": "object",
                        "description": "Identifies whether the paper includes a link to the code associated with the dataset.",
                        "properties": {
                            "Has Code": {
                                "type": "boolean",
                                "description": "Set to true if all code related to data collection, preprocessing, and generation is publicly available in an accessible repository (e.g., GitHub)."
                            },
                            "reference": {
                                "type": "string",
                                "description": "The location in the paper where the code is described (e.g., 'Section 4.1')."
                            },
                            "reasoning": {
                                "type": "string",
                                "description": "Justification based on the paper content explaining why the code is said to be available or not in the paper."
                            }
                        },
                        "required": ["Has Code", "reference", "reasoning"],
                        "additionalProperties": false
                    },
                    "documentation": {
                        "type": "object",
                        "description": "Determines whether the paper provides documentation on the dataset creation process.",
                        "properties": {
                            "Has Documentation": {
                                "type": "boolean",
                                "description": "Set to true if the dataset creation process is documented in the paper."
                            },
                            "reference": {
                                "type": "string",
                                "description": "Where in the paper the process is described (e.g., 'Section 4.1')."
                            },
                            "reasoning": {
                                "type": "string",
                                "description": "Justification based on paper content for the documentation status."
                            }
                        },
                        "required": ["Has Documentation", "reference", "reasoning"],
                        "additionalProperties": false
                    }
                },
                "required": ["code", "documentation"],
                "additionalProperties": false
            }
        },
        "required": ["reproducibility"],
        "additionalProperties": false
    },
    "strict": true
}
  
    
\end{lstlisting}
\end{tcolorbox}

\section{Benchmark Filtering Template}
\label{section:appendix-benchmark-classifier-template}

\begin{tcolorbox}[colback=gray!10,colframe=black,title=Benchmark Filtering Template,breakable]
\#\#\# TASK\\
Given the title and abstract of a paper, determine whether the paper introduces a new dataset. Respond with "true" if the paper introduces a new dataset; otherwise, respond with "false". \\

\#\#\# INPUT \\
Title: \\
\{\{ title \}\} \\

Abstract: \\
\{\{ abstract \}\} \\

\#\#\# OUTPUT FORMAT \\
Return a JSON response in the following format:\\
\{\{\\
    "explanation": "Very short explanation why the answer is true or false",\\
    "answer": "Final boolean answer between true or false"\\
\}\}\\

\#\#\# RESPONSE
\end{tcolorbox}

\end{document}